\documentclass[lettersize,journal]{IEEEtran}
\usepackage{amsmath,amsfonts}
\usepackage{amssymb}
\usepackage{algorithmic}
\usepackage{algorithm}
\usepackage{array}
\usepackage[caption=false,font=normalsize,labelfont=sf,textfont=sf]{subfig}
\usepackage{textcomp}
\usepackage{booktabs} 
\usepackage{stfloats}
\usepackage{url}
\usepackage{multirow}
\usepackage{bm}
\usepackage{threeparttable}  
\usepackage[justification=centering]{caption} 
\usepackage[colorlinks=true, citecolor=blue, linkcolor=blue, urlcolor=red]{hyperref}
\usepackage{verbatim}
\usepackage{tabularx} % 在导言区添加
\usepackage{graphicx}
\usepackage{textcomp}
\usepackage{cite}
\usepackage{pifont}
\begin{document}

\title{E2E-Fly: An Integrated Training-to-Deployment System for End-to-End Quadrotor Autonomy}

% VisFactory, VisCross
%\title{UniFly: A Comprehensive End-to-End Learning-based Policy Training Framework for Quadrotors bridging the Sim-to-Real Gap}

%\title{UniFly: A Integrated Framework for End-to-End Quadrotor Autonomy}

\author{Fangyu Sun\dag, Fanxing Li\dag, Linzuo Zhang, Yu Hu, Renbiao Jin, Shuyu Wu, Wenxian Yu*, Danping Zou*
% \thanks{This work was supported by the National Key Research and Development Program of China (2022YFB3903801) and the National Science Foundation of China (62073214).}
\thanks{All authors are with the School of Automation and Perception, Shanghai Jiao Tong University, China (e-mail: dpzou@sjtu.edu.cn).$^{(*)}$ denotes the corresponding author.{(\dag)} denotes equal contribution.}}

\markboth{}%
{Shell \MakeLowercase{\textit{et al.}} A Comprehensive End-to-End Vision-based Policy Training Framework for Quadrotors for bridging the Sim-to-Real Gap}

% Remember, if you use this, you must call \IEEEpubidadjcol in the second

\maketitle

% \begin{abstract}
%     Transferring learning-based methods from simulation to the real world for quadrotors remains a challenging problem. This is due to several challenges: slow rendering speed of vision-based simulation, gaps in real dynamics and system delays, gaps between real and simulated physics, discrepancies caused by real noise, and the lack of an efficient and comprehensive platform for validating the deployment of algorithms from sim-to-real. Although current research can accomplish many learning-based quadrotor end-to-end control tasks, few platforms enable the full workflow for zero-shot deployment in the real world, making algorithm implementation and reproducibility difficult. To address these issues, we propose UniFly, an end-to-end quadrotor deployment platform from simulation to the real world. We build a complete workflow architecture that enables training from scratch to real-world in a few minutes, based on the optimized high-speed rendering simulator VisFly, sim-to-sim and hardware-in-the-loop validation interfaces, and betaflight-ctrl, which bridges high-level policy decisions and low-level commands.
% We also propose the first comprehensive manual for reward functions and curriculum designs for learning-based quadrotor tasks, as well as a complete alignment from sim-to-real, including system identification, domain randomization, latency compensation and noise interference. Finally, we validated the effectiveness of our UniFly system through a diverse set of benchmark deployment experiments. 
% \end{abstract}

\begin{abstract}

Training and transferring learning-based policies for quadrotors from simulation to reality remains challenging due to inefficient visual rendering, physical modeling inaccuracies, unmodeled sensor discrepancies, and the absence of a unified platform integrating differentiable physics learning into end-to-end training. While recent work has demonstrated various end-to-end quadrotor control tasks, few systems provide a systematic, zero-shot transfer pipeline, hindering reproducibility and real-world deployment. To bridge this gap, we introduce E2E-Fly, an integrated framework featuring an agile quadrotor platform coupled with a full-stack training, validation, and deployment workflow. The training framework incorporates a high-performance simulator with support for differentiable physics learning and reinforcement learning, alongside structured reward design tailored to common quadrotor tasks. We further introduce a two-stage validation strategy using sim-to-sim transfer and hardware-in-the-loop testing, and deploy policies onto two physical quadrotor platforms via a dedicated low-level control interface and a comprehensive sim-to-real alignment methodology, encompassing system identification, domain randomization, latency compensation, and noise modeling. To the best of our knowledge, this is the first work to systematically unify differentiable physical learning with training, validation, and real-world deployment for quadrotors. Finally, we demonstrate the effectiveness of our framework for training six end-to-end control tasks and deploy them in the real world.
\end{abstract}


\section{Introduction} 

End-to-end learning has gained increasing attention in quadrotor autonomy by directly mapping raw sensor observations to actuator commands through neural networks \cite{nav1,nav2,af1,af2}. Unlike conventional modular architectures, which decompose the whole task into sequential modules, such as perception, planning, and motor control, end-to-end policies bypass those intermediate computations, thereby reducing system latency and mitigating error propagation across modules. This data-driven paradigm has demonstrated compelling performance in agile flight \cite{af1,af2}, obstacle avoidance \cite{ob2,ob3}, and high-speed trajectory tracking \cite{track3,track5}, outperforming traditional model-based controllers in highly dynamic or uncertain environments. Within this paradigm, differentiable physics learning, which leverages analytical gradients of the quadrotor dynamics for efficient policy training, has demonstrated impressive performance in quadrotor tasks such as hovering \cite{hover} and high-speed collision avoidance \cite{yuang,vi1}. By providing exact gradients, it substantially improves learning efficiency and final policy quality compared with black-box reinforcement learning (RL) and standard imitation learning approaches.

Despite recent successes, the practical application of end-to-end methods to real-world quadrotors is still challenging. Firstly, training an end-to-end control policy involves choosing appropriate simulators and learning algorithms as well as designing effective rewards for the given task. Secondly, quick and safe validation must be particularly required for quadrotor tasks, as crashes can result in severe damage not only for the vehicle but also for the onboard sensors and computers. Finally, accurate sim-to-real alignment is essential for real-world deployment, including system identification, sensory noise modeling, and control latency compensation. All those factors motivate the development of a system that integrates simulation, learning algorithms, and hardware into a cohesive training-to-deployment pipeline. Such a unified system can substantially accelerate the train-to-deployment procedure, improving the efficiency and safety during debugging and evaluating the trained networks. 

\begin{table*}[htbp]
\centering
\caption{Key characters comparison of state-of-the-art approaches in this area.}
\label{table:previous_work}
\begin{tabular}{ccccccccc}
\toprule
 & \textbf{Type} & \textbf{Hardware} & \textbf{Render} & \textbf{Sensors} &\textbf{Algorithm} & \textbf{Validation}  &\textbf{Deployment} & \textbf{Number of Tasks}\\
\midrule
\textbf{Flightmare\cite{flightmare}} & simulator & $\times$ & Unity & D,R,S,I & RL &HIL &$\times$ &$\times$\\
\textbf{RotorS\cite{rotors}} & simulator & $\times$ & OpenGL & D,R,I & $\times$& $\times$  &$\times$ & $\times$\\
\textbf{CrazyS\cite{crazys}} & simulator & $\times$ & OpenGL & D,R,I & $\times$ & $\times$ &$\times$ & $\times$\\
\textbf{CrazySim\cite{crazysim}} & simulator & $\times$ & OpenGL & D,R,S,I & $\times$ & $\times$&$\times$ & $\times$ \\
\textbf{FlightGoggles\cite{flightgoggles}} & simulator & $\times$ & Unity & D,R,S,I & $\times$ & HIL&$\times$ & $\times$\\
\textbf{FastSim\cite{fastsim}} & simulator & $\times$ & Unity & D,R,S,I,L & RL & HIL &$\times$ & $\times$\\
\textbf{AirSim\cite{airsim}} & simulator & $\times$ & Unreal & D,R,S,I& $\times$& HIL  &$\times$& $\times$\\
\textbf{PyBulletDrone\cite{pybullet}} & simulator & $\times$ & OpenGL & D,R,S,I& RL & $\times$ &$\times$ & $\times$\\
\textbf{OmniDrone\cite{omnidrone}} & simulator & $\times$ & Isaac Sim & D,R,S,I,L & RL& $\times$ &$\times$ & $\times$\\
\textbf{VisFly\cite{visfly}} & simulator & $\times$ & Habitat-sim & D,R,S,I & RL,DS&$\times$ &$\times$  & $\times$\\
\textbf{Agilicious\cite{agilicious}} & platform  & $\checkmark$ & Unity & D,R,S,I & $\times$& HIL &LC & $\times$\\
\textbf{AirGym\cite{airgym}} & platform  & $\checkmark$ & Isaac Sim & D,R,I & RL& $\times$  &SI,DR & 5\\
\midrule
\textbf{E2E-Fly} & platform & $\checkmark$ & Habitat-sim & D,R,S,I & RL,DS &HIL,sim-to-sim  &SI,LC,DR,NM & 6\\
\bottomrule
\end{tabular}
\label{system_com}
\begin{tablenotes}
  \item[a] I, R, D, S, and L refer to IMU, RGB, depth, segmentation, and LiDAR, respectively. HIL refers to hardware-in-the-loop validation. RL and DS represent reinforcement learning and differentiable simulation. SI, LC, DR, and NM represent system identification, latency compensation, domain randomization, and noise modeling, respectively. $\checkmark$ indicates support, $\times$ indicates no support. E2E-Fly is a system-level quadrotor platform that provides the entire technical stack from training to validation and to deployment, including a customization parallel training environment, two hardware platforms for offboard and onboard validation, a fast renderer based on Habitat-sim, four types of sensors, sim-to-sim, and hardware-in-the-loop
  interfaces, algorithms covering RL and differentiable simulation, and four sim-to-real alignment techniques.
\end{tablenotes}
\end{table*}

Many research platforms for quadrotors have been developed, while only a few of them support training-to-deployment effectively. Agilicious \cite{agilicious} is a co-designed hardware and software research platform for aggressive autonomous flight, featuring a high thrust-to-weight ratio (5g acceleration) and a powerful onboard GPU-based computer (Nvidia Jetson TX). Its modular software stack, including a rendering-based simulator and hardware-in-the-loop support, provides flexibility for both traditional perception–planning–control pipelines and learning-based methods. However, its primary focus lies in enabling high-performance flight rather than offering a thorough workflow for end-to-end policy training or systematic sim-to-real transfer.
AirGym \cite{system_thu}, in contrast, is specifically designed for RL. It integrates an RL training environment with hardware for validation, employing PX4 for low-level control and an RK3588-based onboard computer. It investigates several practical considerations for successful sim-to-real transfer, such as system identification, action smoothing, and domain randomization. Nevertheless, AirGym does not incorporate differentiable physical learning algorithms and lacks guidance on reward design for common quadrotor tasks and evaluation on different learning algorithms. 

While these systems are valuable, there remains a need for a unified framework that integrates advanced simulators, state-of-the-art learning algorithms, comprehensive reward design, and training guidelines for common tasks, and detailed protocols for validation and sim-to-real alignment. Such a system would significantly lower the barrier for developing robust end-to-end quadrotor controllers and provide a consistent methodology for both researchers and practitioners.

To this end, we present E2E-Fly, a unified training-to-deployment pipeline designed specifically for developing end-to-end control policies for a diverse range of quadrotor tasks. E2E-Fly integrates advanced simulators, state-of-the-art learning algorithms, and comprehensive guidelines for reward design, system identification, and sim-to-real alignment. The platform also provides two reference quadrotor designs for real-world evaluation, together with six representative benchmark tasks that illustrate the full workflow of training, validating, and deploying end-to-end policies using algorithms such as PPO \cite{ppo} and Back-Propagation Through Time (BPTT) \cite{bptt}.

For the training stage, E2E-Fly adopts VisFly\cite{visfly}, a high-performance differentiable simulator capable of rendering multi-modal sensory observations at high frame rates while providing end-to-end differentiability through the quadrotor dynamics. This enables the use of highly efficient gradient-based learning techniques such as BPTT\cite{bptt}, offering substantially improved sample efficiency compared with black-box RL such as PPO\cite{ppo}. To further support policy learning across diverse tasks, we provide structured reward-design guidelines rooted in common principles of quadrotor flight, including progress-based shaping, smooth action penalization, attitude or velocity regularization, and collision-avoidance costs. We also include curriculum-learning examples for obstacle avoidance and racing, demonstrating how progressive task shaping improves convergence and robustness.

Before real-world deployment, E2E-Fly offers two validation stages: sim-to-sim and hardware-in-the-loop. For sim-to-sim validation, we extend the AirSim \cite{airsim} simulator with streamlined, Betaflight-compatible interfaces that expose all core control channels while removing redundant operations. For hardware-in-the-loop validation, E2E-Fly is integrated with a motion-capture system so that real-time pose information from the physical quadrotor drives the simulator, which in turn generates virtual sensory input—allowing the end-to-end policy to be executed on a real quadrotor in an obstacle-free environment while interacting with virtual obstacles. This significantly reduces the risk of damaging hardware during early-stage evaluation.

Finally, to support seamless deployment, we design and release two quadrotor hardware platforms tailored for policies trained with E2E-Fly. Building on these platforms, we develop a four-stage sim-to-real alignment methodology encompassing system identification, latency compensation, domain randomization, and sensor-noise modeling, with the supporting \textit{betaflight-ctrl} package enabling robust communication and low-level control implementation. The complete pipeline is extensively validated on six benchmark tasks across both hardware platforms.

In summary, E2E-Fly is the first system to unify differentiable physical simulation, first-order gradient-based learning, structured validation, and real-world deployment into a cohesive workflow for quadrotor end-to-end control. To better understand our contributions, we compare E2E-Fly with previous systems in Table \ref{table:previous_work}. By providing an integrated stack spanning software and hardware, the platform enables both rapid policy development and reliable physical execution, thereby accelerating research iteration cycles and lowering the barrier to applying end-to-end methods in agile quadrotor flight.

% \begin{figure*}[htbp]
%     \centering
%     \includegraphics[width=1.0\linewidth]{pic/Overviewfinal2.pdf}
%     \caption{\textbf{The overview architecture of E2E-Fly.}}
%     \label{fig:system}
% \end{figure*}

\section{Related Work}
% We provide a systematic comparison of representative state-of-the-art approaches and the proposed framework, focusing on key technical components and system integration. The results, summarized in Table \ref{system_com}, show that our framework constitutes a unified, comprehensive platform that covers the entire policy development pipeline, from training to validation to real-world deployment.

\subsection{Simulators and Platforms for Quadrotors}

The development of quadrotor simulators and platforms has remarkably accelerated policy training. Existing systems can be broadly categorized into simulators and integrated software-hardware platforms.

Among simulators, Gazebo \cite{gazebo} provides a versatile 3D simulation foundation with a powerful physics engine, upon which RotorS \cite{rotors} offers modular pipelines for verifying state estimators and controllers. Specialized tools like CrazyS \cite{crazys} and CrazySim \cite{crazysim} deliver software-in-the-loop tests for Crazyflie drones. To enhance visual realism, several simulators, including FlightGoggles \cite{flightgoggles}, Flightmare \cite{flightmare}, FastSim \cite{fastsim}, and AirSim \cite{airsim}, leverage commercial game engines like Unity and Unreal. PyBulletDrone \cite{pybullet} and OmniDrone \cite{omnidrone} prioritize physics computation speed for parallel RL training. VisFly \cite{visfly}, built on Habitat-Sim \cite{habitat}, further supports high-speed rendering, differentiable simulation, and end-to-end policy training. In terms of integrated software-hardware platforms, Agilicious \cite{agilicious} provides an open-source, co-designed quadrotor platform by integrating high-performance computing units with agile flight hardware and offering a modular software stack. AirGym \cite{airgym} integrates hardware, a simulator, and a deployment chain to validate RL-based sim-to-real transfer on physical systems.

Collectively, these contributions have significantly advanced the field. However, they exhibit three limitations: (1) a focus on either simulation fidelity or hardware integration, but rarely both, (2) insufficient support for differentiable physical learning within a unified workflow, and (3) fragmented or under-specified sim-to-real alignment techniques. Consequently, the community still lacks a fully integrated platform that enables systematic sim-to-real transfer, supports diverse learning paradigms, and facilitates zero-shot policy deployment across various tasks.

\begin{figure*}[htbp]
    \centering
    \includegraphics[width=1.0\linewidth]{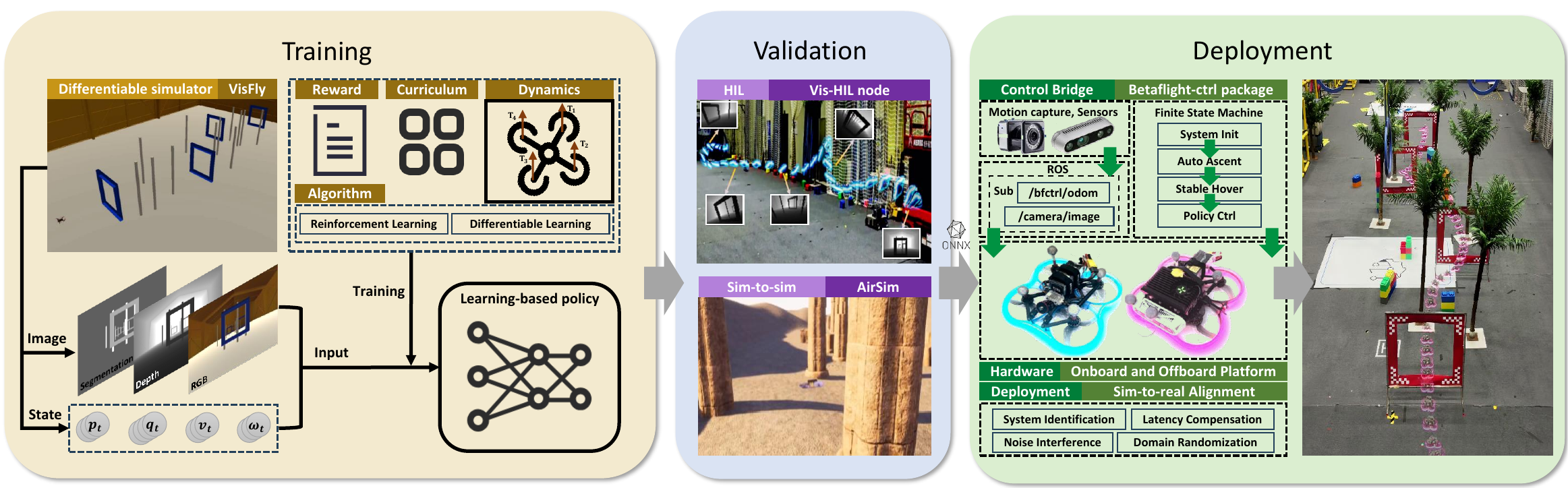}
    \caption{\textbf{The overview architecture of E2E-Fly.} In the training phase, the state-based and vision-based inputs are acquired from VisFly. During this process, the reward function is designed according to the reward function manual, while the accurate dynamics model supports training both via RL and differentiable simulation. The trained policy can be directly transferred to AirSim via an internal interface for cross-platform sim-to-sim test. In the real-world deployment phase, we provide two quadrotor hardware platforms that support real-time hardware-in-the-loop simulation and onboard flight test, respectively. The transfer from policy outputs to low-level control commands is achieved through our open-source \textit{betaflight-ctrl} package.}
    \label{fig:system}
\end{figure*}

\subsection{Reward Function Design for Quadrotors}%%% ！！！
% A well-designed reward function can not only improve training efficiency but also make the policy more robust and smooth-behaved when deployed on real drones. 
The design of the reward function critically governs the performance of learned policies by defining the solution space of the optimization problem. Despite the diversity of quadrotor tasks, effective reward structures often share key insights about stability, agility, and perceptual awareness.

Traditional learning-based approaches often decompose rewards into sparse and dense formulations. Sparse rewards, such as binary success-failure signals, offer straightforward design principles but suffer from low sample efficiency, challenging exploration requirements, and inherent non-differentiability \cite{sparse1,sparse2}. In contrast, dense rewards provide incremental guidance by penalizing state deviations or incentivizing task progress \cite{progress1,progress2}, while maintaining temporal consistency and differentiability. For most quadrotor control tasks, sparse and dense rewards can be used together. Zhao et al. \cite{ob1} employ a dense velocity-tracking reward alongside a sparse penalty for hazards and errors to guide the agent in adaptive flight through obstacle-aware environments. Xie et al. \cite{gap} use a dense position reward together with a sparse gap-angle constraint to guide the drone through a narrow gap. In autonomous drone racing \cite{progress1, progress2, ra1,ra2}, training typically employs a high-weighted progress reward, a substantial crash penalty, and a sparse success reward to encourage rapid traversal of the gate while avoiding collisions. Some studies also leverage dense–sparse reward combinations to tackle complex dynamic tasks, such as traversing swinging circles \cite{cir1,cir2}, flying through dynamic gaps \cite{dong_gap1}, and performing dynamic obstacle avoidance \cite{yuang}. 

%Despite these task-specific successes, the field lacks a systematic, general-purpose guidelines for reward design. Current approaches rely on manual tuning and expert intuition, highlighting the need for a unified methodology that captures shared insights across diverse quadrotor missions.

Despite these isolated achievements, no systematic guideline for reward design exists. Current approaches rely heavily on heuristic tuning and expert knowledge, indicating a clear need for a structured methodology that abstracts common insights across a wide range of quadrotor tasks.

% Despite the success of task-specific reward functions, a systematic framework for general-purpose reward design across diverse quadrotor missions remains lacking.

\subsection{Sim-to-Real Alignment for Quadrotors}
Sim-to-real refers to the process of safely and efficiently transferring algorithms developed in simulation to the real world. Eliminating the gaps between simulated and real environments is the main challenge for algorithm deployment. These gaps mainly stem from the dynamic discrepancies, latency, and observation noises that are difficult to calibrate. 

% In recent years, researchers have proposed numerous approaches to reduce the sim-to-real gap. Among these, domain randomization \cite{domain1,domain2,domain3,domain4,domain5,domain6} has been empirically established as the most effective trick for deploying policies learned in simulation to real robots. Randomly varying physical parameters and sensor noise in simulation compel deep networks to learn features that are robust to sim-to-real discrepancies, enabling zero-shot transfer without costly real-world data collection. In addition, several system-identification-based alignment approaches \cite{agilicious,sysID2,sysID3} have been proposed to compensate for the dynamic discrepancies. These methods effectively mitigate dynamics-induced differences by measuring the actual system parameters and incorporating them into the simulation. To address system noise introduced by visual inputs, researchers \cite{vi1,vi2,vi3} have proposed distinct solutions for optical flow, RGB images, and depth maps, respectively. Except that,  Dionigi et al. \cite{input1} investigate how observation-space design in quadrotor control policies affects zero-shot sim-to-real transfer, filling the gap in analyzing the impact of input configuration on policy performance. Chen et al. \cite{matter} discuss the factors that influence quadrotor sim-to-real transfer from network inputs, reward design, domain randomization, and system identification, but only touch on a few aspects in each, without providing a comprehensive solution.

In recent years, numerous approaches have been introduced to narrow the sim-to-real gap. Among these, domain randomization \cite{domain1,domain2,domain3,domain4,domain5,domain6} has been empirically shown to be one of the most effective techniques. Randomly varying physical parameters and adding sensor noise during training encourages the deep network to learn robust features that generalize across sim-to-real discrepancies. In parallel, system identification \cite{agilicious,sysID2,sysID3} has been developed to compensate for dynamic inaccuracies and latency. These approaches incorporate measured real-world system parameters into the simulation, effectively reducing dynamics-induced mismatches. To address noise introduced by visual inputs, researchers propose specialized solutions tailored to optical flow \cite{vi1}, RGB image \cite{vi3}, and depth map \cite{vi2}, respectively. Beyond these, Dionigi et al. \cite{input1} investigate how the design of the observation space in quadrotor control policies influences sim-to-real transfer, thereby filling a gap in understanding the role of input. Chen et al. \cite{matter} identify several points that influence quadrotor policies transfer, including training randomization and appropriate parameter identification. Building upon these insights, we further establish a comprehensive framework for minimizing the sim-to-real gap by combining system identification, latency compensation, domain randomization, and noise modeling into a complete and detailed solution.

\section{System Overview}
The architecture of our E2E-Fly is illustrated in Fig. \ref{fig:system}. During the training phase, the policy receives image and state observations from VisFly, supported by reward design, accurate and differentiable quadrotor dynamics, and curriculum learning, thereby enabling training through both RL and differentiable physical learning algorithms. In the validation phase, the trained policy can be directly evaluated through the simulator's internal interfaces, including sim-to-sim transfer and hardware-in-the-loop test. For real-world deployment, sim-to-real alignment is achieved through system identification, latency compensation, domain randomization, and noise modeling. Finally, the policy is deployed onto the hardware platform, where the \textit{betaflight-ctrl} package converts policy commands into low-level control signals, thereby completing an end-to-end pipeline for policy training, validation, and deployment.

The structure of this paper is as follows. Sec. \ref{training} details the training pipeline, including the simulator, learning-based algorithms, reward design formats for quadrotor tasks, and curriculum learning. Sec. \ref{validation} demonstrates the policy validation framework, covering both sim-to-sim transfer and hardware-in-the-loop simulation. Sec. \ref{deployment} covers real-world deployment, introducing the hardware platform, low-level control bridge, and four concrete alignment techniques for sim-to-real transfer. Sec. \ref{exp} validates the effectiveness of the proposed system through six comprehensive benchmark experiments. Sec. \ref{dis} discusses the comparative advantages and disadvantages of RL and differentiable simulation, along with strategies for combining their strengths. 

\section{Training}\label{training}
This section focuses on the training methodology, detailing the simulator, learning-based policy training, general-purpose reward design for quadrotor tasks, and curriculum learning.

\subsection{Fast Rendering Simulator: VisFly}
VisFly \cite{visfly} is a versatile quadrotor simulator that enables parallel multi-agent training with synchronized multi-modal perception (RGB, depth, segmentation) and high-frequency state/IMU data. Leveraging Habitat-sim's rendering engine \cite{habitat}, it achieves up to 6000 FPS at $256\times256$ depth resolution. Built on a fully differentiable dynamics model implemented in PyTorch, it supports analytical gradients and GPU-accelerated physics, facilitating differentiable physics algorithms. The framework provides interfaces for RL and differentiable simulation, enabling efficient parallel training for advanced quadrotor tasks. It incorporates four established control interfaces: single-rotor thrust, mass-normalized collective thrust and body rates (CTBR), position and yaw, and linear velocity and yaw. In the subsequent section, we primarily employ the CTBR as the policy output, as it has been proven \cite{CTBR1,CTBR2,CTBR3} to be one of the optimal low-level control commands for a quadrotor’s end-to-end tasks.

% First, we have added the capability of custom multi-scene updating for complex environments. Although VisFly already supports parallel training for multiple agents, previous versions only allowed multiple agents to train and update in the same scene. Now, it is possible to allocate different numbers of agents to different scenes for parallel updates. For example, 100 agents can be divided into 4 groups across 4 different scenes, with 25 agents training in each scene. This update method is essential for complex tasks, such as racing in cluttered environments, as it allows agents to learn more varied scene information more quickly. 

% Building on the existing capabilities, this paper further optimizes VisFly. The original VisFly is solely used for policy training in simulation and does not provide a sim-to-real interface. In this paper, by integrating sim-to-sim, the \textit{betaflight-ctrl} package, and the hardware-in-the-loop (HIL) node within VisFly, we have successfully achieved zero-shot deployment of VisFly's policy within minutes using \texttt{rostopic}. We not only provide a sim-to-real validation process but also offer sim-to-sim and HIL simulation guidance for vision-based policies. Through system identification and real-world experiments, we have thoroughly addressed the challenge of zero-shot transfer of learning- and vision-based policies from simulation to the real world. The specific implementations of \textit{betaflight-ctrl} package, sim-to-sim, and HIL will be detailed in Sec. \ref{bf}, Sec. \ref{sim-to-sim}, and Sec. \ref{hil}.

The initial version of VisFly is limited to simulation-based training, whereas our enhanced framework integrates sim-to-sim and hardware-in-the-loop validation, and a \textit{betaflight-ctrl} interface. We have further integrated a comprehensive set of sim-to-real alignment techniques into VisFly, including system identification, latency compensation, domain randomization, and noise modeling. These additions support zero-shot policy deployment in minutes via \texttt{rostopic}, providing complete sim-to-real workflows for learning-based policies. The specific implementations of \textit{betaflight-ctrl} package, sim-to-sim, hardware-in-the-loop simulation, and sim-to-real alignment will be detailed in Sec. \ref{bf}, Sec. \ref{sim-to-sim}, Sec. \ref{hil}, and Sec. \ref{sim-to-real}.

\subsection{Policy Training Algorithm}
\subsubsection{Reinforcement Learning}

We formulate the quadrotor control task as an infinite-horizon Markov Decision Process (MDP), defined by the tuple $(\mathcal{S}, \mathcal{A}, \mathcal{P}, \mathcal{R}, \gamma)$. The state space $\mathcal{S}$ and the action space $\mathcal{A}$ are continuous. The state transition function $\mathcal{P}:\mathcal{S}\times\mathcal{A}\rightarrow\mathcal{S}$, describes the system dynamics, determining the next state via $s_{t+1} = \mathcal{P}(s_t, a_t)$. The reward function $\mathcal{R}: \mathcal{S} \times \mathcal{A} \rightarrow \mathbb{R}$ assigns an immediate reward $r_t = \mathcal{R}(s_t, a_t)$ at each time step. The objective function is defined in Eq. \ref{return}.
\begin{equation}\label{return}
\mathcal{J}(\theta) = \mathbb{E}_{s_0 \sim \rho_0, a_t \sim \pi_{\theta}(\cdot|s_t)}\left[ \sum_{t=0}^{\infty} \gamma^{t} \mathcal{R}(s_t, a_t) \right]
\end{equation}
where $\mathcal{J}(\theta)$ represents the expected discounted return, $s_0$ is sampled from the initial state distribution $\rho_0$, $s_t \in \mathcal{S}$, $a_t \in \mathcal{A}$, and $\gamma \in [0, 1)$ denotes the discount factor.
The goal is to learn an optimal policy $\pi_{\theta}^{*}$ with parameters $\theta$ that maximizes $\mathcal{J}(\theta)$ from the initial state distribution. The optimization objective is thus defined in Eq. \ref{rl}.
\begin{equation}\label{rl}
\pi_{\theta}^{*} = \operatorname*{argmax}_{\pi_{\theta}} \mathcal{J}(\theta)
\end{equation}

When the policy is optimized via gradient ascent, the parameters are updated according to Eq. \ref{gradient_ascent}.
\begin{equation}\label{gradient_ascent}
\theta_{t+1} \leftarrow \theta_t + \alpha{\nabla_\theta \mathcal{J}(\theta)}
\end{equation}
where $\alpha$ is the learning rate, ${\nabla_\theta \mathcal{J}(\theta)}$ represents an estimate of the policy gradient. If the reward function and dynamics model are differentiable, we can compute an accurate gradient via differentiable simulation.

\subsubsection{Learning via Differentiable Simulation}
BPTT \cite{bptt} leverages first-order gradients within differentiable simulation, enabling precise computation of the gradient as expressed in Eq. \ref{ds1}. 
\begin{equation}\label{ds1}
    \nabla_\theta \mathcal{J}(\theta) = \left( \sum_{t=0}^{N-1} \gamma^t \frac{\partial \mathcal{R}(s_t, a_t)}{\partial \theta} \right)
\end{equation}
where $N$ represents the length of the horizon. The analytic first-order gradient can be computed via backward chain propagation, as shown in Eq. \ref{ds2}.
\begin{equation}\label{ds2}
\begin{aligned}
    &\nabla_\theta \mathcal{J}(\theta) = \\
    &\sum_{t=0}^{N-1} \gamma^t  \sum_{i=1}^t \frac{\partial \mathcal{R}(s_t, a_t)}{\partial s_t} \prod_{j=i}^t \left( \frac{\partial s_j}{\partial s_{j-1}} \right) \frac{\partial s_i}{\partial a_i}\frac{\partial a_i}{\partial \theta}   \\
    &+ \sum_{t=0}^{N-1} \gamma^t \frac{\partial \mathcal{R}(s_t, a_t)}{\partial a_t} \frac{\partial a_t}{\partial \theta}
\end{aligned}
\end{equation}

For quadrotors, the transition function is governed by the differential dynamics $\dot{ s_t}=\mathcal{P}(s_t,a_t)$. In practical implementation, this continuous process is discretized into a time-stepped evolution of the system, expressed as $s_{t+1} = s_t + \mathcal{P}(s_t,a_t)\,dt$.

Compared with the estimated gradient in Eq. \ref{return}, BPTT achieves faster convergence and higher sample efficiency by using analytic gradient, making it particularly well-suited for continuous control tasks in robotics with accurate dynamics and differentiable rewards.

\subsubsection{Quadrotor Dynamics}\label{dynamic}
We employ a fully differentiable quadrotor dynamics \cite{visfly} as the state transition function $\dot{s_t}$ in Eq. \ref{dynamic}. 
\begin{equation}\label{dynamic}
    \dot{{s}_t} = 
    \begin{bmatrix}
    \dot{\bm{p}}_t^{W} \\
    \dot{\bm{q}_t} \\
    \dot{\bm{v}}_t^{W} \\
    \dot{\bm{\omega}_t}
    \end{bmatrix}
    =
    \begin{bmatrix}
    \bm{v}_t^{W} \\
    \bm{q}_t \cdot 
    \begin{bmatrix}
    0 \\
    \bm{\omega}_t/2
    \end{bmatrix} \\
    \frac{1}{m} \left( \bm{R}_t^{WB} \left( \bm{f}_t^{T} + \bm{f}_t^{D} \right) \right) + \bm{g} \\
    \bm{J}^{-1} \left( \bm{\tau}_t^{T} + \bm{\tau}_t^{D} - \bm{\omega}_t \times \bm{J} \bm{\omega}_t \right) \\
    \end{bmatrix}
\end{equation}
where $m$ and $\bm {J}$ are mass and diagonal moment of inertia matrix, $\bm {p}_t^W$ and $\bm {v}_t^W$ are the position and velocity vector in the world frame, unit quaternions $\bm q_t$ with $||\bm q_t ||=1$ are used to represent orientations, $\bm{R}_t^{WB}$ is the rotation matrix from the body frame to the world frame, $\boldsymbol{\omega}_t$ represents angular velocity in the body frame. $\bm {g}=[0,0,-9.81]^T$ denotes the gravity vector. $\bm {f}_t^T$ and $\bm\tau_t^T$ denote the collective thrust in the body-z axis and cumulative body torques produced by the rotors, as shown in Eq. \ref{f_thrust_and_tau_thrust}. 
\begin{equation}\label{f_thrust_and_tau_thrust}
    \bm f_t^{{T}} = \sum_i^{4} \bm f_t^i , \qquad
    \bm \tau_t^{{T}} = \sum_i^{4} (\bm\tau_t^i + \bm r^{i} \times \bm f_t^i)
\end{equation}
where $\bm f_t^i$ and $\bm \tau_t^i$ are the thrust and torque generated by motor $i$, and $\bm r^{i}$ is the arm length of the force. 

We employ quadratic functions to model the relationship between the thrust and torque generated by a single motor and its rotational speed $\Omega_t$, as shown in Eq. \ref{quadratic}.
\begin{equation}\label{quadratic}
\bm f_t^i = \begin{bmatrix}
k_{f0} \\
k_{f1}\cdot \Omega_t \\
k_{f2} \cdot \Omega_t^2
\end{bmatrix},
\quad \bm\tau_t^i = 
\begin{bmatrix}
k_{\tau 0} \\
k_{\tau 1} \cdot \Omega_t \\
k_{\tau 2} \cdot \Omega_t^2
\end{bmatrix}
\end{equation}
where $k_{f0}$,$k_{f1}$, $k_{f2}$ and $k_{\tau0}$,$k_{\tau1}$, $k_{\tau2}$ represent the parameters for the two quadratic functions, respectively. The motor is simulated using a first-order system. The dynamic equation for each motor is defined in Eq. \ref{motor}.
\begin{equation}\label{motor}
    \dot{\Omega}_t = \frac{1}{k^{\mathrm{motor}}} (\Omega_t^{\mathrm{cmd}} - \Omega_t)
\end{equation}
where $k^{\mathrm{motor}}$ is the motor time constant, $\Omega_t^{\mathrm{cmd}}$ is the commanded rotor speed. Air drag force $\bm f_t^D$ and torque $\bm\tau_t^D$ are also modeled for aggressive motion. It is worth noting that our estimation of drag is precise to both the first-order and second-order drag coefficients shown in Eq. \ref{fd}.

\begin{equation}\label{fd}
    \bm{f}_t^D =    
    % -k_1\begin{bmatrix}
    % 0 \\
    % {\bm{v}}_t^{B} \\
    % 0
    % \end{bmatrix}-
        \bm{k}_D \circ
    % \begin{bmatrix}
    % 0 \\
    % 0 \\
    % ({{\bm{v}}_t^{B}})^2
    % \end{bmatrix}
    {{\bm{v}}_t^{B}} \circ {{\bm{v}}_t^{B}}
    % k_1\bm {v}_B-k_2||\bm {v}_B||\bm {v}_B
\end{equation} 
where ${\bm{v}}_t^{B}$ represents the velocity in the body frame, $\bm{k}_D$ is the coefficient of second-order drag simulating the friction drag at high speeds. The simulation of the drag component is crucial for our system to achieve stability at low speeds and aggressiveness at high speeds.

\subsection{Reward Design for Quadrotor Tasks}\label{reward_mannul}
The efficacy of learning-based controllers is fundamentally governed by the reward function, which shapes the policy's behavior by defining the optimization landscape that guides its actions. To establish a systematic methodology, this section presents a comprehensive reward design framework encompassing both dense and sparse formulations, with all parameters $\lambda_i$ assumed positive unless specified otherwise. It should be noted that the reward functions in this section are denoted directly without unified normalization. 

\textbf{Dense Rewards:}
Dense rewards provide continuous, differentiable learning signals essential for policy optimization, particularly when leveraging BPTT via differentiable simulation. These formulations enable stable gradient-based learning across the state-action space.

\paragraph{\textbf{Progress}} 
% The fundamental objective for a quadrotor in any task is to progress toward a desired target. In traditional paradigms, this is abstracted as waypoint tracking or the execution of motion primitives. Conversely, in learning-based methods, this core objective is typically formulated as a progress reward, which motivates the policy network to acquire the same navigational behavior. The most popular progress reward \cite{progress2} is defined in Eq. \ref{prog1}.

The fundamental objective for a quadrotor in most tasks is to progress toward a desired target. In traditional paradigms, this objective is typically achieved through waypoint tracking or the execution of motion primitives. In learning-based frameworks, the same behavior is formulated as a progress reward, which motivates the policy to move toward the goal. A widely adopted formulation for this reward \cite{progress2} is given by Eq. \ref{prog1}.

\begin{equation}\label{prog1}
        r_{t}^{\text{prog1}} = \lambda_1 \cdot \left\|  d_{t-1} -  d_t \right\|   
\end{equation}
This reward promotes swift flight toward the goal by incentivizing a reduction in distance to the target. We usually use a significant positive parameter $\lambda_1$ to make it a leading term during aggressive tasks. Where $d_t$ represents the distance between the drone's position $\bm p_t$ and goal position $\bm g_t$ defined in Eq. \ref{dt}. 
\begin{equation}\label{dt}
        d_t = \left\| \bm p_t- \bm g_t \right\|
\end{equation}
More directly, the agent is rewarded for reducing the distance to the goal, thereby promoting proximity to the target, as formulated in Eq. \ref{prog2}.
\begin{equation}\label{prog2}
        r_{t}^{\text{prog2}} = -\lambda_2 \cdot \left\| d_t  \right\|  
\end{equation}
For tasks with stability requirements like hovering and landing, it's beneficial to decouple rewards for the xy-plane and z-axis displacements. This aligns with control architectures that require independent steering. Accordingly, we design a planar progress reward $r_{t}^{\text{progxy}}$ (Eq. \ref{prog3}), alongside individual axis-wise rewards $r_{t}^{\text{progx}}, r_{t}^{\text{progy}}, r_{t}^{\text{progz}}$ (Eqs. \ref{prog4}-\ref{prog6}) for precise independent control.
\begin{equation}\label{prog3}
        r_{t}^{\text{progxy}} = -\lambda_3 \cdot \left\| d_t^{xy} \right\|  
\end{equation}
\begin{equation}\label{prog4}
        r_{t}^{\text{progx}} = -\lambda_4 \cdot \left\|  d_t^{x} \right\|   
\end{equation}
\begin{equation}\label{prog5}
        r_{t}^{\text{progy}} = -\lambda_5 \cdot \left\|  d_t^{y} \right\|   
\end{equation}
\begin{equation}\label{prog6}
        r_{t}^{\text{progz}} = -\lambda_6 \cdot \left\|  d_t^{z} \right\|   
\end{equation}
\paragraph{\textbf{Smooth Action}} A penalty on action changes is imperative for smooth control. This practice, similar to minimizing control derivatives in traditional optimization. With a normalized action $\bm{u}$, such a penalty typically takes the form of an action smoothness cost, as exemplified in Eqs. \ref{act1} and \ref{act2}.
\begin{equation}\label{act1}
    r_{t}^{\text{act1}} = \left\| \bm{u}_{t} \right\|
\end{equation}
\begin{equation}\label{act2}
        r_{t}^{\text{act2}} = \left\| \bm{u}_{t} - \boldsymbol{u}_{t-1} \right\|
\end{equation}
where $\bm u_t$ represents the policy output, $r_{t}^{\text{act1}}$ and $r_{t}^{\text{act2}}$ are used to punish the current action and two consecutive actions, respectively. We use Eq. \ref{act3} to combine these two rewards to encourage a smooth trajectory. 
\begin{equation}\label{act3}
        r_{t}^{\text{act3}} = -(\lambda_7\cdot r_{t}^{\text{act1}}+\lambda_8\cdot r_{t}^{\text{act2}})
\end{equation}

\paragraph{\textbf{Orientation}}
The quadrotor's orientation is critical in perception-aware tasks \cite{ob1,track5,nav2}, because it directly influences perception quality and mission success. In these scenarios, simply tracking a positional trajectory is inadequate. The orientation must be independently planned to maintain optimal target visibility, ensure feature observability, and satisfy dynamic constraints. This requirement is inherently supported by the differential flatness property \cite{diff-flatness,minimum-snap} of quadrotor dynamics. Consequently, a common method for encouraging orientation alignment is to employ a yaw angle-based error, as defined in Eq. \ref{ori1}.
\begin{equation}\label{ori1}
    r_{t}^{\text{ori1}} = \lambda_9 \cdot \exp(-\psi_{t}^{e})
\end{equation}
where the $\psi_{t}^{e}$ is the relative yaw angle between the quadrotor and target. An exponential decay is chosen to ensure high sensitivity to angle variations. Another way is to penalize the quaternion error to align the orientation directly, as shown in Eq. \ref{ori2}.
\begin{equation}\label{ori2}
    r_{t}^{\text{ori2}} = -\lambda_{10} \cdot \left\|\bm q_{t} - \bm q_{t}^{des}\right\| 
\end{equation}
where $\bm q_{t}^{des}$ is the desired orientation of the next target.

In \cite{ori4}, the author uses a perception-aware reward to maximize the visibility of the next goal, as shown in Eq. \ref{ori3}.
\begin{equation}\label{ori3}
    r_{t}^{\text{ori3}} =\lambda_{11}\cdot\exp(-\left\|\bm u_t^2+\bm v_t^2+\bm{\dot u}_t^2+\bm{\dot v}_t^2)\right\|)
\end{equation} 
where $(\bm u_t,\bm v_t)$ and $(\bm{\dot u_t},\bm{\dot v_t})$ represent the pixel position and velocity of the next target in the image plane of the camera. 

In specialized maneuvers, such as traversing a narrow gap, the quadrotor must align its roll angle with the gap. The corresponding reward that enforces this specific alignment is formulated in Eq. \ref{ori4}.
\begin{equation}\label{ori4}
        r_{t}^{\text{ori4}} = 
        \begin{cases}
        -\min\left(\tan\left|\phi_t^e\right|, \lambda_{12}\right) & \text{approaching gap} \\
        0 & \text{otherwise}
        \end{cases}
\end{equation}
where $\phi_t^e$ is the relative roll angle between the quadrotor and the gap.

\paragraph{\textbf{Linear Velocity}} 
Maintaining controllable flight in perception-aware and stability-critical tasks necessitates explicit constraints on the quadrotor's linear velocity. We provide two approaches: the first directly penalizes linear velocity to suppress aggressive motion, as formalized in Equation \ref{linearv1}; the second employs reference-guided principles to enforce adherence to a predefined speed profile through reward shaping, implemented via Equations \ref{linearv2} and \ref{linearv3}.

\begin{equation}\label{linearv1}
    r_{t}^{\text{v1}} = -\lambda_{13} \left\| \bm{v}_{t} \right\| 
\end{equation}
\begin{equation}\label{linearv2}
    r_{t}^{\text{v2}} = -\lambda_{14}\cdot (\left\| \bm{v}_{t} \right\| - v_{d})
\end{equation}
\begin{equation}\label{linearv3}
    r_{t}^{\text{v3}} = -\lambda_{15}\cdot (\exp(\left\| \bm{v}_{t} \right\| - v_{d})+1)
\end{equation}
where the scalar $v_{d}$ denotes the desired speed magnitude. Unlike the linear penalty in Eq. \ref{linearv2}, Eq. \ref{linearv3} introduces an exponential term that grows rapidly when the velocity exceeds $v_d$. By introducing a $+1$ offset, a baseline penalty is maintained even when the target velocity is not exceeded. This asymmetric response effectively suppresses aggressive flight tendencies, proving particularly beneficial for scenarios requiring stable operation.

High-stability tasks, including hovering and landing, require independent regulation of vertical velocity, coordinated with the quadrotor's height to ensure a safe, controlled approach to the ground. This specific functional dependency is formulated in Eq. \ref{linearv4}.
\begin{equation}\label{linearv4}
    r_{t}^{\text{vz}} = \lambda_{16}\cdot \frac{1}{1+| {v}_{t}^h + v_t^{z} |}
\end{equation}
where the $ v_t^{z}$ is the z-axis velocity of the quadrotor. The desired adaptive landing velocity $\bm {v}_{t}^h$ is defined in Eq. \ref{linearvz}.
\begin{equation}\label{linearvz}
    v_{t}^h = (\alpha \cdot  d_t^z).\operatorname{clamp}(v^{min},v^{max})
\end{equation}

where $\alpha$ is the velocity coefficient, $ d_t^z$ denotes the height of the quadrotor above the ground, $v^{max}$ and $v^{min}$ represent the maximum and minimum descent velocity, respectively.
    
\paragraph{\textbf{Angle Velocity}}
To suppress high-frequency oscillations and ensure stable attitude control, a penalty on angular velocity is recommended. This formulation is especially pertinent when the policy generates CTBR commands, as it directly regularizes the actuation signal, implemented in Eq. \ref{anglev}.
\begin{equation}\label{anglev}
    r_{t}^{\text{anglev}} = -\lambda_{17} \left\| \bm{\omega}_{t} \right\|
\end{equation}
        
\paragraph{\textbf{Collision Avoidance}} 
Collision avoidance is fundamental to autonomous drone navigation in cluttered or dynamic environments. To this end, the reward structure must incorporate mechanisms that actively discourage proximity to obstacles and other agents. We thus formulate collision-avoidance reward terms that penalize unsafe distances, promoting evasive maneuvers before a possible collision. Among these, Eq.~\ref{avoid1} adopts an inverse-distance formulation, providing a computationally efficient and interpretable means of enforcing safety margins.
\begin{equation}\label{avoid1}
    r_{t}^{\text{avoid1}} = \lambda_{18} \left(\frac{1}{d_t^{col}+b}\right) 
\end{equation}
where $ d_t^{col}$ represents the distance to the nearest obstacle, and $b$ is a small positive constant to prevent division by zero. Eq. \ref{avoid2} presents a more refined formulation \cite{yuang}, integrating both distance-based and velocity-dependent terms:

\begin{equation}\label{avoid2}
\begin{aligned}
    r_{t}^{\text{avoid2}} = & - \left\|\bm v_t^c \right\|\max\left(1 - ( d_t^{col} - r_q), 0\right)^2 \\
    & + \lambda_{19} \ln(1 + \exp ({\lambda_{20} ( d_t^{col} - r_q)}))
\end{aligned}
\end{equation}
where $\bm v_t^c$ is the approach velocity toward the nearest obstacle and $r_q$ is the quadrotor's radius. The reward combines a truncated quadratic potential and a soft-plus barrier, scaled by approach velocity. This velocity scaling ensures the penalty is active only during approach, with the penalty increasing as the approach speed increases to promote timely deceleration.

% \begin{equation}\label{avoid3}
%     r_{\text{col3}} = \lambda_{17}\cdot\frac{{p}_{\text{target}} \cdot {p}_{\text{ego}}}{\|{p}_{\text{target}}\|_2 \cdot \|{p}_{\text{ego}}\|_2}  \\
% \end{equation}
    
% These collision avoidance rewards are crucial for ensuring the safe operation of drones, especially in tasks that involve high speeds or complex environments with multiple obstacles. By integrating these penalties into the overall reward function, we can guide the drone's policy to avoid collisions while still optimizing for task completion.

\textbf{Sparse Reward:} To guide the agent toward successful task completion, sparse rewards are provided upon reaching each target, while penalties are imposed for crashes. These event-driven signals are inherently sparse and non-differentiable, serving as critical terminal feedback for policy optimization.

\paragraph{\textbf{Goal and Crash}}
The sparse rewards for achieving the goal and for crashes are shown in Eqs. \ref{goal} and \ref{collision}.
\begin{equation}\label{goal}
    r_{t}^{\text{goal}} = 
    \begin{cases}
    \lambda_{21} & \text{arrive the goal} \\
    0 & \text{otherwise}
    \end{cases}
\end{equation}

\begin{equation}\label{collision}
    r_{t}^{\text{crash}} = 
    \begin{cases}
    -\lambda_{22} & \text{collision occur} \\
    0 & \text{otherwise}
    \end{cases}
\end{equation}

The experiments in Sec. \ref{exp} employ the reward design methodology outlined in this section to formulate each task-specific reward and accomplish the tasks. Table \ref{tab:reward_tasks} summarizes the proposed reward function formulations for key foundational tasks considered in this work.

\begin{table}[htbp]
\centering 
\caption{The reward design instruction for different types of tasks.}
\label{tab:reward_tasks} 
\scriptsize
\begin{tabular}{llll} 
\toprule 
\textbf{Alg.} & \textbf{Task} &\textbf{Reward} \\ \midrule
\multirow{5}{*}{PPO} &Hovering & $r_{t}^{\text{prog}}+r_{t}^{\text{act}}+r_{t}^{\text{v}}+r_{t}^{\text{anglev}}+r_{t}^{\text{goal}}+r_{t}^{\text{crash}}$ \\ 
&  Landing&$r_{t}^{\text{prog}}+r_{t}^{\text{act}}+r_{t}^{\text{v}}+r_{t}^{\text{anglev}}+r_{t}^{\text{goal}}+r_{t}^{\text{crash}}$ \\
& Tracking &$r_{t}^{\text{prog}}+r_{t}^{\text{act}}+r_{t}^{\text{ori}}+r_{t}^{\text{v}}+r_{t}^{\text{anglev}}+r_{t}^{\text{goal}}+r_{t}^{\text{crash}}$ \\
&Racing &$r_{t}^{\text{prog}}+r_{t}^{\text{act}}+r_{t}^{\text{goal}}+r_{t}^{\text{crash}}$ \\
& Obstacle avoidance &$r_{t}^{\text{prog}}+r_{t}^{\text{ori}}+r_{t}^{\text{v}}+r_{t}^{\text{anglev}}+r_{t}^{\text{avoid}}+r_{t}^{\text{goal}}+r_{t}^{\text{crash}}$ \\
\midrule
\multirow{5}{*}{BPTT} &Hovering & $r_{t}^{\text{prog}}+r_{t}^{\text{act}}+r_{t}^{\text{ori}}+r_{t}^{\text{v}}+r_{t}^{\text{anglev}}$   \\
&  Landing&$r_{t}^{\text{prog}}+r_{t}^{\text{act}}+r_{t}^{\text{v}}+r_{t}^{\text{anglev}}$ \\
& Tracking &$r_{t}^{\text{prog}}+r_{t}^{\text{act}}+r_{t}^{\text{ori}}+r_{t}^{\text{v}}+r_{t}^{\text{anglev}}$  \\ 
&Racing&$r_{t}^{\text{prog}}+r_{t}^{\text{anglev}}+r_{t}^{\text{v}}$  \\
&Obstacle avoidance&$r_{t}^{\text{prog}}+r_{t}^{\text{ori}}+r_{t}^{\text{v}}+r_{t}^{\text{anglev}}+r_{t}^{\text{avoid}}$  \\
\bottomrule 
\end{tabular}
\begin{tablenotes}
  \item[a] The table above provides the reward function designs for all experiments in this paper, with each reward term derived from the reward function manual in the methodology section. This design instruction serves as the standard for the subsequent experimental section.
\end{tablenotes}
\end{table}

\subsection{Curriculum Learning}
% For relatively simple tasks such as hovering, tracking, landing, and racing, we adopt a one-step training approach, as these tasks are generally tractable for learning-based methods. However, for more complex scenarios, such as fast obstacle avoidance or racing in cluttered environments, this approach often leads the agent to converge to local optima. To mitigate this issue, we incorporate a structured curriculum learning that enables the agent to acquire the skills for task completion progressively.

% For instance, in high-speed, collision-free flight through complex environments, the training process begins with simple point-to-point navigation in an obstacle-free environment. The environmental complexity and flight speed are then gradually increased through curriculum stages. We emphasize that, given a well-designed curriculum and appropriate parameterization, variations in training environments do not significantly impair the agent’s ability to master each sub-task or the final objective. To illustrate the role of curriculum learning, Table~\ref{table:curriculum} summarizes the curriculum design schemes for representative complex tasks.
Based on task complexity, we employ distinct training strategies. For fundamental tasks such as hovering and landing, a direct one-step training paradigm proves effective. In contrast, complex scenarios such as high-speed obstacle avoidance or navigation in densely cluttered environments often cause agents to converge to local optima under one-step training. To address this limitation, we implement a structured curriculum learning strategy that progressively guides the agent through increasingly challenging sub-tasks.

For example, training for high-speed collision-free flight in complex environments begins with simple point-to-point navigation without obstacles. Through sequential curriculum stages, both environmental complexity and flight speed are systematically increased. We demonstrate that with appropriate curriculum design and parameterization, such environmental variations do not hinder the agent's ability to master individual sub-tasks or the overall objective. Table~\ref{table:curriculum} summarizes specific curriculum designs for representative complex tasks, illustrating this progressive learning approach.

\begin{table}[htbp]
\centering
\caption{The examples of curriculum designs for complex tasks.}
\label{table:curriculum}
\begin{tabular}{ll}
\toprule
\textbf{Task} & \textbf{Curriculum Design} \\
\midrule
\multirow{3}{*}{\textbf{High speed obstacle avoidance}} & C1: obstacle-free point navigation\\ 
& C2:  sparse obstacle navigation\\
& C3:  high-speed collision-free flight\\
\midrule
\multirow{3}{*}{\textbf{Racing with random obstacles}} & C1: obstacle-free racing\\
& C2: sparse obstacle racing \\
& C3: dense obstacle racing \\
\bottomrule
\end{tabular}
\begin{tablenotes}
  \item[a] C1, C2, and C3 represent curriculum 1, curriculum 2, and curriculum 3, respectively.
\end{tablenotes}
\end{table}
\section{Validation}\label{validation}
This section presents two complementary strategies for policy validation: cross-simulator (sim-to-sim) transfer and hardware-in-the-loop simulation.
\subsection{Sim-to-sim: From VisFly to Airsim}\label{sim-to-sim}
Direct transfer of control policies trained in a single simulator to physical quadrotors poses significant risks, particularly for aggressive maneuvers where unavoidable sim-to-real gaps in aerodynamics, system dynamics, latency, and sensor noises can critically compromise performance. To mitigate these challenges, sim-to-sim validation serves as an intermediate step, enabling verification of policy logic consistency, assessment of generalization capability under varied dynamics, and identification of simulator-specific biases before real-world deployment. This progressive verification strategy substantially enhances policy robustness while systematically reducing the sim-to-real gap.

We implement a cross-simulator validation framework that transfers policies trained in VisFly to AirSim before real-world deployment. AirSim \cite{airsim}, an open-source simulator built on Unreal Engine, offers high-fidelity physics and photorealistic sensor simulation. Our framework establishes a direct interface between VisFly and AirSim, enabling policy evaluation under identical control logic across simulators. We further develop an AirSim-compatible version of the \textit{betaflight-ctrl} package that maintains all core control interfaces while eliminating redundant remote operations. This integration allows policies to be validated in AirSim without modification, ensuring consistency throughout the sim-to-real pipeline. 

\subsection{Hardware-in-the-loop simulation: \textit{VIS-HIL} node}\label{hil} 
Hardware-in-the-loop simulation integrates physical quadrotor hardware into a simulated visual environment, forming a real-time closed-loop system that enables rigorous validation of control and perception algorithms under realistic yet safe conditions. This approach is particularly critical for perception-aware agile flight, where real-world testing not only imposes safety risks but also exposes vulnerable onboard sensors and computing units to potential damage during aggressive maneuvers. To address these challenges, the E2E-Fly framework implements this paradigm by operating a physical quadrotor within a motion-capture system while rendering photorealistic virtual scenes, building upon prior work \cite{flightgoggles}. Unlike purely synthetic simulations, this approach preserves the authentic dynamics and proprioceptive sensing of the real vehicle, while facilitating testing in arbitrarily complex and visually rich scenarios—all without incurring physical collision risks \cite{agilicious}. Consequently, it significantly accelerates the development cycle for vision-based agile flight systems by mitigating the safety and cost constraints inherent in physical testing.

\begin{figure}[htbp]
    \centering
    \includegraphics[width=1.0\linewidth]{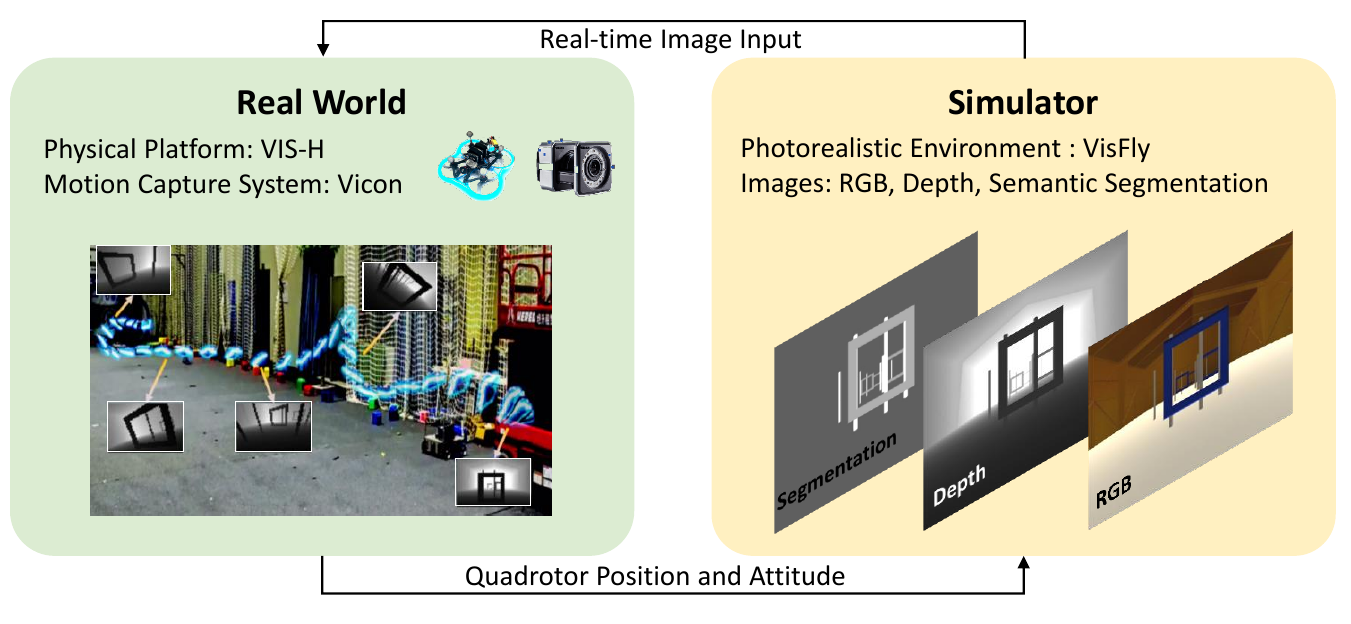}
    \caption{\textbf{The hardware-in-the-loop simulation in E2E-Fly.} It consists of a real quadrotor flying in a motion capture system combined with a photorealistic simulation of complex 3D environments. Multiple sensors can be simulated with minimal delays while virtually flying in various simulated scenes. Such hardware-in-the-loop simulation offers a modular framework for prototyping robust vision-based algorithms safely, efficiently, and inexpensively.}
    \label{fig:drones}
\end{figure}

To implement hardware-in-the-loop verification within E2E-Fly, we developed a dedicated simulation node, \textit{VIS-HIL}, which synchronizes the real quadrotor with its simulated counterpart. The node subscribes to the quadrotor’s real-time pose via \texttt{rostopics}, aligns it with the corresponding coordinate frame in VisFly, and generates simulated visual observations that match the physical pose. These visual data, published at a configurable frequency ($30 Hz$ by default), serve as inputs for policy inference, while the quadrotor’s state and dynamics are measured directly from the motion capture system and IMU. The \textit{VIS-HIL} node supports multiple visual modalities, including RGB, depth, and semantic segmentation, enabling flexible evaluation of perception-based control policies under realistic and controlled conditions.

\section{Deployment}\label{deployment}
In this section, we present a comprehensive methodology for deploying trained policies onto physical quadrotors, encompassing two distinct hardware platforms, a low-level control bridge, and a complete four-stage sim-to-real alignment framework.
\subsection{Quadrotor Platform}
We design two quadrotor platforms, VIS-Real (VIS-R) and VIS-Hardware-in-the-loop (VIS-H), for both onboard and offboard real-world experiments. The structures are based on frame OddityRC, illustrated in Fig. \ref{fig:drones}. 

\begin{figure}[htbp]
    \centering
    \includegraphics[width=1.0\linewidth]{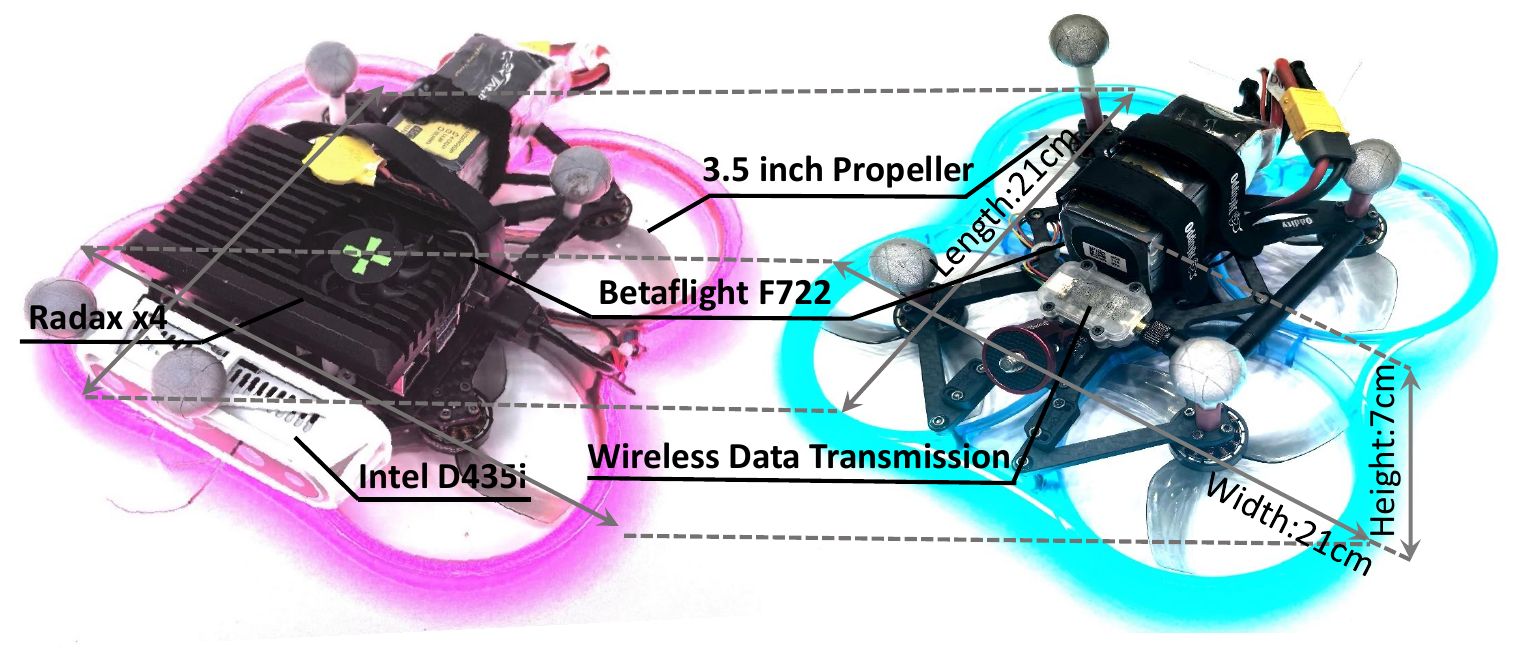}
    \caption{\textbf{Proposed quadrotor platforms: VIS-R and VIS-H.} VIS-R is used for onboard experiments with a Radax X4 and an Intel D435i RGB-D camera. VIS-H supports offboard and hardware-in-the-loop experiments via a wireless data transmitter and a wireless video transmitter.}
    \label{fig:drones}
\end{figure}

% The VIS-R is equipped with the Radax X4 as its onboard computer, featuring an Intel N100 processor that can perform real-time inference for lightweight neural networks. Additionally, VIS-R features a front-mounted Intel RealSense D435i stereo depth camera, enabling simultaneous RGB-D data acquisition. In contrast to the VIS-R, the VIS-H employs a wireless data transmitter, a wireless video transmitter, and an FPV camera for real-time command execution and image transmission via remote ground-station computation. This offboard processing paradigm enhances operational safety during initial test of aggressive tasks by preventing damage to the onboard computational hardware in the event of a crash. Moreover, based on the VIS-H offboard verification paradigm, we can leverage remote ground station computing to perform real-time inference for large-scale models by a data transmitter.

The VIS-R integrates a Radax X4 onboard computer with an Intel N100 processor, capable of executing real-time inference for lightweight neural networks. It is further equipped with a front-facing Intel RealSense D435i stereo camera for synchronized RGB-D perception. In contrast, the VIS-H adopts an offboard processing architecture, utilizing a wireless data link and an FPV camera to stream images and receive control commands from a ground station in real time. This design enhances operational safety during initial tests of aggressive maneuvers by isolating high-value computing hardware from potential crash damage. Furthermore, the VIS-H’s communication architecture enables its use in hardware-in-the-loop experiments, where it can interface with a high-fidelity simulator while leveraging ground-station resources to execute larger-scale models in real time.

To enhance the agility and maximum speed, the masses of VIS-R and VIS-H are 750g and 470g (excluding the battery). The flight controller utilizes the Betaflight-compatible STM32F7-based Mini flight controller, paired with 3000KV motors and 3.5-inch D90s propellers. Comprehensive system identification parameters, including detailed electromechanical specifications, are provided in Sec. \ref{systemID}). The overall cost of the VIS-R and VIS-H systems is approximately 630\$ and 280\$, respectively.

\subsection{Low-Level Control Bridge: \textit{betaflight-ctrl}}\label{bf}
% The betaflight-ctrl package has two primary functions. First, it employs a finite-state machine (FSM) to switch in and out of the trained policy command as needed. Second, it can fit the hover throttle value, which is crucial for the drone's hover mode. Our FSM comprises six states: INIT, AUTO\_TAKEOFF, AUTO\_HOVER, AUTO\_LAND, CMD\_TAKEOFF, and CMD\_CTRL, representing initialization, automatic takeoff, automatic hover, automatic landing, command-controlled takeoff, and command-controlled mode, respectively.

% Before each launch of the offboard program, the FSM will first enter the AUTO\_TAKEOFF mode to take off automatically at a preset altitude. Afterward, the throttle stick should be moved to the middle position to enter the AUTO\_HOVER mode. At this point, the drone can be manually controlled to the initial point for program execution (which is generally related to domain randomization). Then, the program can be started, and by moving the gear on the remote controller, the drone will enter the CMD\_CTRL mode, where it will fly according to the policy outputs. If the drone deviates from the planned trajectory or encounters a hazardous situation, the AUTO\_HOVER mode can be automatically activated by switching the gear button or by moving the pitch/roll stick (the right stick if using left-hand throttle). In conclusion, the betaflight-ctrl package effectively mitigates the risks associated with program malfunctions.

The \textit{betaflight-ctrl} package integrates a finite-state machine (FSM) with four primary operational modes: System Initialization, Autonomous Ascent, and Stabilized Hover, as well as a Policy Control mode for executing policy commands. The FSM first autonomously elevates the drone to a predefined altitude, then stabilizes it in hover. From this state, the quadrotor can either be manually repositioned or switched into Policy Control, in which the drone follows the outputs of the trained policy. To ensure operational safety, the FSM enables rapid fallback to Stabilized Hover via manual input or automatic fallback when deviations or hazards are detected, thereby safeguarding against failures of the learned policy or offboard program.

Meanwhile, the \textit{betaflight-ctrl} package establishes communication between the onboard computer and the flight controller, enabling real-time publishing and utilization of telemetry data (including attitude, acceleration, battery voltage, remote control signals, and flight modes), thereby facilitating experiment analysis. Moreover, \textit{betaflight-ctrl} incorporates an adaptive throttle–acceleration mapping, supports diverse command modes (position, velocity–altitude, acceleration, attitude–thrust, CTBR), and provides auxiliary tools for diagnostics and development, including PID analysis, CPU load monitoring, and automated \texttt{rosbag} logging. 

\subsection{Sim-to-real Alignment}\label{sim-to-real}
\subsubsection{\textbf{System Identification}}\label{systemID}
% System identification minimizes the sim-to-real gap, ensuring that control policies designed are more effective when deployed in the real world, especially in aggressive tasks. The design of the end-to-end policy relies on an accurate description of the system model. In this paper, we conducted a comprehensive system identification for sim-to-real transfer, which not only involves precise measurement of quadrotor dynamics parameters but also complete latency compensation and alignment of the response curves between the simulation and the real world. We detail our system identification process in this section, along with the specific reasons for conducting these alignments. Given its significance, latency compensation will be elaborated in a separate section in \ref{latency}.

System identification plays a critical role in narrowing the sim-to-real gap, particularly for aggressive flight tasks where model inaccuracy can severely compromise policy effectiveness. Since the design of an end-to-end policy fundamentally depends on an accurate system representation, we perform a comprehensive identification process that encompasses not only the precise measurement of physical parameters but also full latency compensation and response alignment between simulation and real-world dynamics. In this section, we detail our system identification methodology and justify the necessity of each alignment step. Given its importance, latency compensation is treated separately in Sec. \ref{latency}.

\begin{figure}[htbp]
    \centering
    \includegraphics[width=1.0\linewidth]{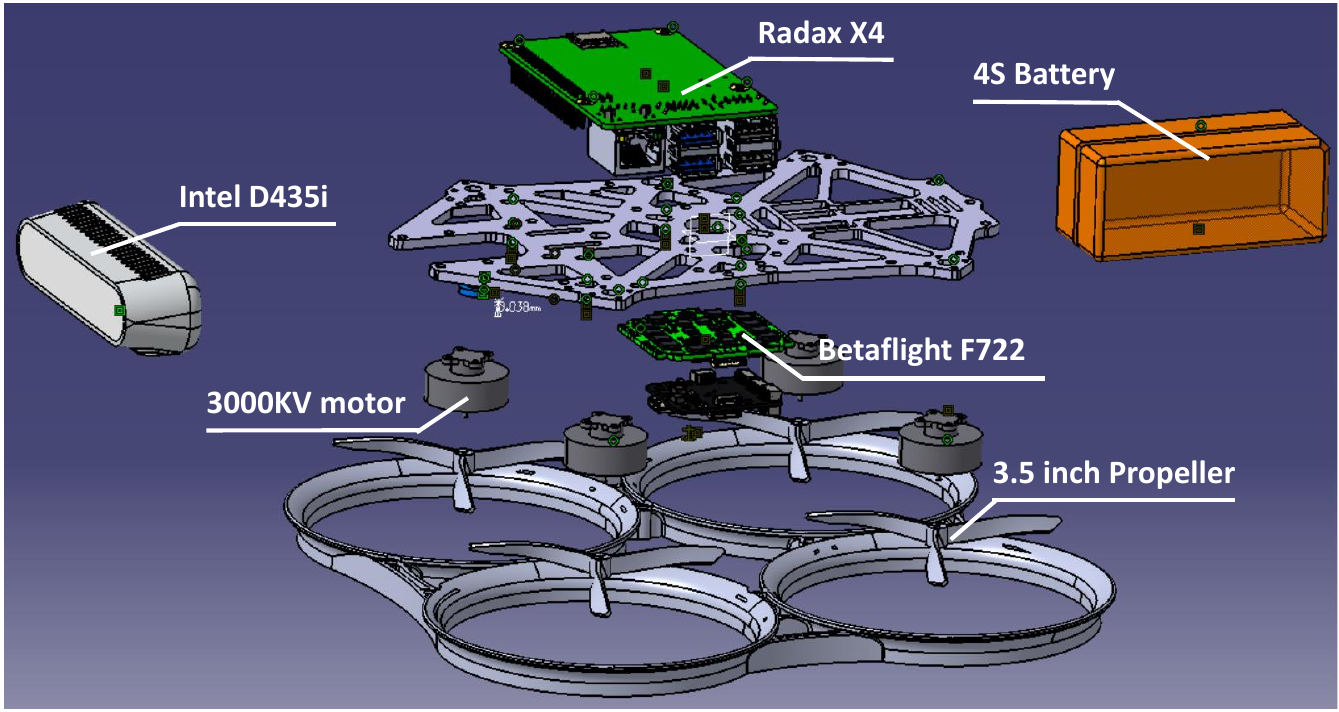}
    \caption{\textbf{The digital prototype employed for moment of inertia test.} The model is constructed by accurately assigning the material and mass properties for each component based on the physical quadrotor. Following this, the inertia tensor is computed within the software environment, enabling the extraction of the three principal moments of inertia directly from its diagonal.}
    \label{fig:shuzi}
\end{figure}

% Since UniFly employs an accurate dynamics model of the quadrotor, we first conducted a quantitative analysis of the physical parameters. The mass, arm length, and moment of inertia are all measured using instruments. The inertia matrix is obtained by calculating the moments of inertia about the quadrotor's three axes, with the error in this part of the system identification controlled to within 10\%. The triple pendulum for measuring the moment of inertia is shown in Fig. \ref{tools}. Next, we measure the thrust map from motor speed to thrust and the delay parameter $k^{motor}$ of the motor response. For this, we employ a high-precision static propeller test stand named LY-5KGF, as shown in Fig. \ref{fig:LY-5KGF}. We use the FV2000 motor and 3.5-inch propellers corresponding to VIS-H and VIS-R, and gradually increase the throttle from 0\% to 100\% at a fixed frequency. We select dozens of valid points to fit a quadratic response curve of speed-thrust and use the coefficients of this quadratic function as the thrust map in the simulation. For the delay response, we record data on the throttle rising from 5\% to 95\% and falling from 95\% to 5\% to fit a first-order delay response model and obtain the motor time constant $k^{motor}$. The identified system parameters are shown in Table \ref{table:parameters}.

Since E2E-Fly employs an accurate dynamics model, we begin system identification by precisely measuring physical parameters. While mass $m$ and arm length $\bm r_i$ are directly measured, the moment of inertia $\bm J$ is obtained through a digital prototype that replicates the actual quadrotor's material composition and weight distribution. The inertia matrix is derived from this computational model, with estimation errors controlled within 10\%. The digital prototype used for inertia calculation is shown in Fig. \ref{fig:shuzi}.

\begin{figure}[htbp]
    \centering
    \includegraphics[width=0.9\linewidth]{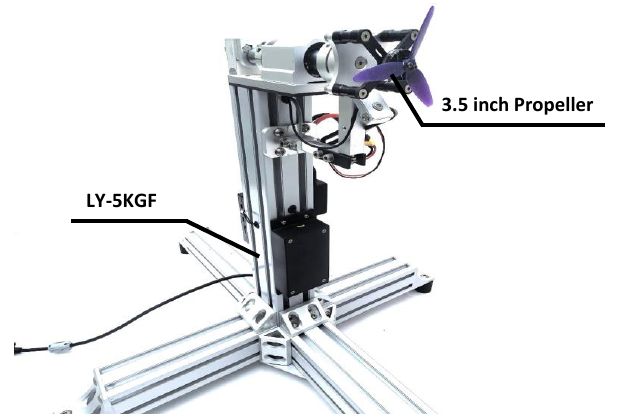}
    \caption{\textbf{The LY-5KGF test stand for motor system identification.} It‘s capable of test propellers of various sizes and featuring software support for remote monitoring and parameter tuning.}
    \label{fig:LY-5KGF}
\end{figure}

We then characterize the motor-propeller system using the high-precision LY-5KGF test stand (Fig. \ref{fig:LY-5KGF}). With FV2000 motors and 3.5-inch propellers matching the VIS-H and VIS-R configurations, we execute a throttle sweep from 0\% to 100\% at fixed intervals. Multiple operating points are sampled to fit a quadratic thrust-speed mapping, whose coefficients are directly embedded in the simulation. For dynamic response identification, we record throttle transitions between 5\% and 95\% to identify a first-order delay model, extracting the motor time constant $k^{motor}$. All identified parameters are summarized in Table \ref{table:parameters}.

\begin{table*}[h]
\centering
\caption{Accurate parameters of our hardware platforms}
\label{table:parameters}
\begin{tabular}{ll}
\toprule
  \textbf{Parameter} & \textbf{Value} \\
\midrule
  mass $m$ [kg] & $0.47$ for VIS-H, $0.75$ for VIS-R \\
  inertia $\bm J$ [kg$\cdot$m$^2$] & [$1.25$, $1.28$, $2.03$]$\times 10^{-3}$ for VIS-H, [$1.41$, $1.53$, $2.05$]$\times 10^{-3}$ for VIS-R \\
  maximum thrust per motor $\bm f^{max}$ [N] & $5.12$ \\
  arm length $\bm r$ [m] & $0.075$ \\
  motor time constant  $k^{motor}$ [s] & $0.035$ \\
  thrust map $[k_{f2},k_{f1},k_{f0}]$ [N$\cdot$s$^2$/rad$^2$, N$\cdot$s/rad, N] & [$4.04\times10^{-7}$, $2.56\times10^{-5}$, $-2.62\times10^{-2}$] \\
  motor drag coefficient $\bm{k}_D$ [N$\cdot$m$\cdot$s/rad] & [$0.05$, $0.05$, $1.15$] \\
  max motor angular velocity $\Omega_{max}$ [rad/s] & $4200$\\
\bottomrule
\end{tabular}
\end{table*}

\subsubsection{\textbf{Latency Compensation}}\label{latency}
% Experimental validation has revealed that latency occurs in two primary forms: motor response latency and system communication latency in the real world. The motor response latency .  We find that the system delay can severely affect sim-to-real performance; failing to account for this delay in simulation can cause oscillations or even loss of control in real-world flight.

All real-world systems with finite computational and communication resources exhibit non-negligible delays, which are further compounded by dynamic response lags and filter-induced phase shifts. The analysis and mitigation of such latency is critical for control performance, particularly when executing policy commands under model mismatch, external disturbances, and actuator constraints. Experimental validation reveals that system latency primarily arises from two sources: motor response latency and system communication latency. The former, corresponding to the inherent dynamics of the system, has been identified and modeled in Sec. ~\ref{systemID}. The latter, if left unmodeled and uncompensated, poses a severe threat to sim-to-real transfer, often inducing high-frequency oscillations or even catastrophic control failure during real-world deployment.

To characterize and compensate for system communication latency, we employ a step-response alignment method. A predefined step signal is applied to the real quadrotor's angular velocity while vertical thrust is maintained at a normalized gravitational acceleration level (suitable for CTBR commands). Real-world step responses for each axis are recorded via \texttt{rosbag}, while an identical input is applied in simulation. Alignment is achieved by introducing initial action-frame delays in the simulated environment and tuning the underlying PID parameters to match both the amplitude and phase of the physical system's response.

Experimental measurements indicate a communication delay of approximately $90 ms$ in telemetry-based offboard operation, reduced to below $ 30 ms$ with onboard computation. Given a control frequency of $ 30 Hz$, this corresponds to 3 and 2 action-frame delays for the VIS-H and VIS-R platforms, respectively. After implementing this compensation, the residual latency discrepancy between simulation and real-world operation becomes negligible, enabling nearly identical closed-loop behavior across both domains.

\subsubsection{\textbf{Domain Randomization}} 
% Many researchers aim to reduce the sim-to-real gap by increasing the variety of domain randomization and the range of noise, enabling agents to learn to handle corner cases they might encounter in the real world. However, this approach undoubtedly increases training time and difficulty. Conversely, insufficient domain randomization can lead to failure in transferring to the real world, as agents may become confused when encountering situations in the real environment that were not seen during simulation training. Therefore, how to properly balance reducing the sim-to-real gap with increasing training complexity through domain randomization is a question worth exploring for quadrotors and the robotics community.
% Building on this foundation, this work provides a set of empirically validated domain-randomization parameters for various tasks, available for researchers to select. These parameters, which primarily comprise the quadrotor's initial setup attributes and task-specific props, are summarized in Table \ref{table:domain_random}. To enhance training efficiency, a targeted randomization configuration is employed for each task. This approach not only improves policy robustness but also effectively balances training speed with generalization capability. Notably, by leveraging our accurate system identification, the domain randomization methodology precludes the need to randomize the quadrotor's physical parameters. This significantly streamlines the policy training process and reduces the complexity of configuring the randomization space.
Many studies attempt to narrow the sim-to-real gap by expanding the diversity of domain randomization and the magnitude of introduced noise, enabling agents to cope with corner cases that may arise in real-world scenarios. However, excessive randomization inevitably increases the training burden and prolongs convergence, whereas insufficient randomization often results in poor transferability, as agents fail to generalize to unseen conditions in the physical environment. Thus, determining an appropriate balance between reducing the sim-to-real gap and maintaining a feasible level of training complexity remains an open challenge for quadrotor learning and the broader robotics community.

Building upon this insight, this work provides a set of empirically validated domain-randomization parameters tailored to various tasks, as summarized in Table~\ref{table:domain_random}. These parameters primarily encompass the quadrotor’s initial configurations and task-specific environmental factors. To enhance training efficiency, a task-oriented randomization scheme is employed, which improves policy robustness while maintaining a balance between learning speed and generalization capability. Furthermore, leveraging accurate system identification, our framework eliminates the need to randomize the quadrotor’s intrinsic physical parameters, thereby simplifying the configuration of the randomization space and streamlining the overall training process.

\begin{table}[htbp]
\centering
\caption{Domain randomization parameters for specific tasks.}
\label{table:domain_random}
\begin{tabular}{lll}
\toprule
\textbf{Task Type} & \textbf{Target Task} & \textbf{Randomization Parameters} \\
\midrule
\multirow{4}{*}{\centering State-based} & Hovering & $\bm p,\bm v$\\
 & Landing & $\bm p$\\
 & Racing & $\bm p$\\
 & Tracking & $\bm p,\bm v$\\
\midrule
\multirow{2}{*}{\centering Vision-based} & Racing with obstacles & $\bm p,\bm p_{\text{gate}},\bm q_{\text{gate}},\bm p_{\text{ob}}$ \\
 & Visual landing & $\bm p,\bm p_{\text{size}},\bm p_{\text{shape}}$ \\ 
\bottomrule
\end{tabular}
\begin{tablenotes}
  \item[a] $\bm p,\bm v,\bm p_{\text{gate}},\bm q_{\text{gate}},\bm p_{\text{ob}},\bm p_{\text{size}},\bm p_{\text{shape}}$ denote the initial position and linear velocity of the quadrotor, the initial position and orientation of the gate, the initial position of the obstacle, and the size and shape of the landing platform, respectively.
\end{tablenotes}
\end{table}

\subsubsection{\textbf{Noise Modeling}}
It is well recognized that part of the sim-to-real gap arises from discrepancies in real-world noise characteristics that are difficult to model in simulation. Interestingly, even without introducing state-based noise to the quadrotor inputs in E2E-Fly, policies trained within our framework demonstrate zero-shot transfer to physical platforms. This behavior underscores the importance of accurate system identification and latency compensation in achieving a precise dynamics model. Nevertheless, such noise-free transferability is primarily effective for state-based quadrotor tasks and remains insufficient for vision-based applications.

For perception-aware tasks such as obstacle avoidance and racing in cluttered environments, E2E-Fly incorporates simulated sensor noise without relying on real-world data. We observe that adding Gaussian noise to the inverse depth map (the reciprocal of depth values) helps narrow the discrepancy between simulated and real-world depth perception. Conceptually, this process can be viewed as directly applying Gaussian domain randomization to the disparity map, thereby improving the robustness of the visual features learned from depth information. To further enhance realism, particularly when emulating sensors such as the Intel D435i stereo RGB-D camera, the Redwood depth noise model \cite{redwoodnoise} can be employed. The detailed configurations of visual and state noise are summarized in Table~\ref{table:noise}.
% As is well known, part of the sim-to-real gap is due to the inability to align real-world noise. However, an interesting phenomenon is that even without introducing state-based noise to the quadrotor (network input) in VisFly, the policy trained by our system still achieves zero-shot transfer. This indirectly highlights the importance of system identification and latency compensation for a precise dynamics model. However, this noise-free zero-shot transfer is only sufficient for state-only quadrotor tasks and is not adequate for vision-based tasks.

% For perception-aware tasks, such as obstacle avoidance and racing in cluttered environments, we simulate the real-world noise in VisFly without using any real-world data. We have found that simply taking the inverse depth map (the reciprocal of the depth values) and adding Gaussian noise can compensate for the sim-to-real depth gap. Of course, if one aims to better reduce the detailed differences caused by images, for example, with the Intel D435i stereo RGB-D camera, we suggest using the redwood depth noise model \cite{redwoodnoise}. The detailed visual and state noise are summarized in Table \ref{table:noise}.
\begin{table}[h]
\centering
\caption{The type of noise provided in E2E-Fly\cite{visfly,habitat}.}
\label{table:noise}
\begin{tabular}{lll}
\toprule
 \textbf{Sensor Type} & \textbf{Classes} \\
 \midrule
 \multirow{5}{*}{\textbf{Vision}} & Gaussian noise &  \\
 & Salt \& pepper noise &  \\ 
 & Poisson noise &  \\
 & Speckle noise &  \\
 & Redwood depth noise \\
\midrule
 \multirow{2}{*}{\textbf{Imu}} & Gaussian white noise &  \\
 & Bias Random Walk & \\
 \bottomrule
\end{tabular}
\end{table}

\begin{figure*}[htbp]
    \centering
    \includegraphics[width=1.0\linewidth]{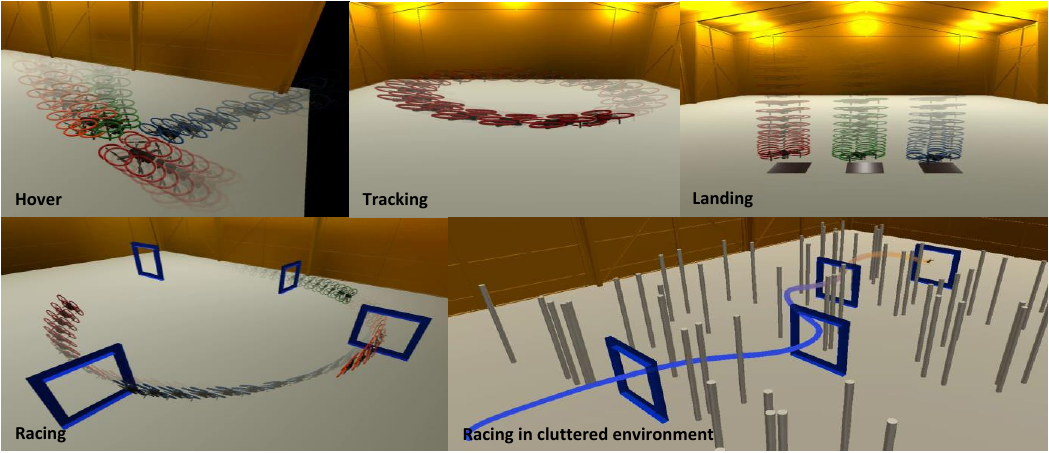}
    \caption{\textbf{Examples of benchmark training scenarios constructed in the E2E-Fly.} We present the simulation performance across various tasks, demonstrating that all policies trained via differentiable simulation and RL can accomplish their objectives.
    }
    \label{fig:task_sim}
\end{figure*}

\section{Experiments and Results}\label{exp}
% \subsection{Observation, Action and Reward}
% We use the standard quadrotor state as the basic observation, which comprises the relative position, attitude, linear velocity, and angular velocity. The relative position, linear velocity and orientation are all transferred to the body axis. Therefore, the basic dimension of proprioceptive observation is 13. State-based observations are tailored to each task by adjusting the underlying state vector. For visual observations, we select the image modality—depth, RGB, or semantic according to the specific task requirement. Leveraging its proven effectiveness in prior end-to-end drone tasks \cite{CTBR1,CTBR2}, we adopt the CTBR command as the action output of our policy. 

\begin{figure*}[htbp]
    \centering
    \includegraphics[width=1.0\linewidth]{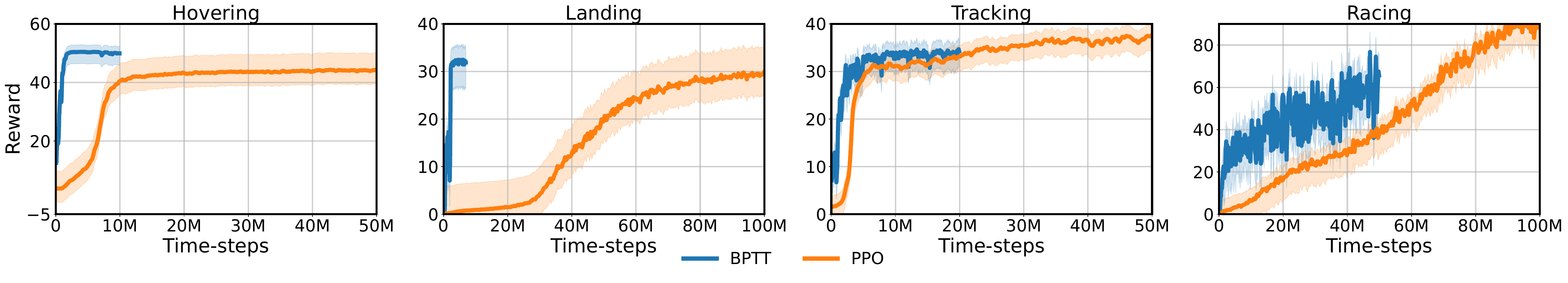}
    \caption{\textbf{Training rewards comparison between PPO and BPTT.} It is evident from the figure that BPTT exhibits faster convergence, higher sample efficiency, and achieves a higher reward compared with PPO.}
    \label{fig:diff_method_reward}
\end{figure*}

\begin{table*}[htbp!]
\centering
\caption{We record the training FPS, the total time steps required to converge, and the overall training time for PPO and BPTT on the four baseline tasks. The FPS is measured with 100 parallel environments.}
\label{table:training}
\begin{tabular}{l*{4}{c}*{4}{c}*{4}{c}}
\toprule
& \multicolumn{4}{c}{FPS [its/s]} & \multicolumn{4}{c}{time steps [its]} & \multicolumn{4}{c}{Time [s]} \\
\cmidrule(lr){2-5} \cmidrule(lr){6-9} \cmidrule(lr){10-13}
 & Hovering & Landing & Tracking & Racing & Hovering & Landing & Tracking & Racing & Hovering & Landing & Tracking & Racing \\
\midrule
\textbf{PPO} & $\approx 20000$ & $\approx 20000 $ & $\approx 20000$ & $\approx 15000$ & 5e7 & 1e8 & 5e7 & 1e8 & $\approx 2500$ & $\approx 5000$ & $\approx 2500$ & $\approx 6600$ \\
\textbf{BPTT} & $\approx13000 $& $\approx13000$ & $\approx24000 $& $\approx20000$ & 1e7 & 1e7 & 2e7 & 5e7 & $\approx 750$ & $\approx 750$ & $\approx 800$ & $\approx 2500$ \\
\bottomrule
\end{tabular}
\end{table*}

\subsection{Benchmark Settings}
\subsubsection{State-based Tasks\cite{ABPT}}
\paragraph{Hovering} 
Hovering, as the most fundamental task, requires the quadrotor to hover stably at the target point from any initial position. In the hovering task, we use a perception-free state  $\bm {s}_t^{\text{drone}} = \left[\bm {p}^{r}_t,\bm {v}_t, \bm {q}_{t}, \boldsymbol{\omega}_{t} \right] \in \mathbb{R}^{13}$ as the observation space, where $\bm {p}^{r}_t$, $\bm {q}_{t}$, $\bm {v}_t$ and $\boldsymbol{\omega}_{t}$ represent the relative position from quadrotor to the target, orientation, linear velocity and angle velocity of the quadrotor at time step $t$ respectively. We define the task error as the Euclidean distance between the agent’s position and the hover target point.
\paragraph{Tracking}
The tracking task requires the quadrotor to start from a random initial point and follow a specified trajectory at a given speed. This task serves as a baseline that bridges traditional and learning-based methods and is particularly important for some RL-based algorithms. The state-only observation space is defined as $\bm {s}_t^{\text{drone}} = \left[\bm {p}^{r_i}_t,\bm {v}_t, \bm {q}_{t}, \boldsymbol{\omega}_{t} \right] \in \mathbb{R}^{10+3i}$, where $\bm {p}^{r_i}_t\in\mathbb{R}^{3i}$ is a concatenation of the relative positions of next $i$ reference points from the quadrotor. In this paper, we set $i=10$. We define the tracking error as the Euclidean distance between the actual trajectory and the target trajectory.
\paragraph{Landing}
The landing task requires the quadrotor to start from a random position, gradually descend, and eventually land at the required position on the ground. The observation space is defined by $\bm {s}_t^{\text{drone}} = \left[\bm {p}^{r}_t,\bm {v}_t, \bm {q}_{t}, \boldsymbol{\omega}_{t} \right] \in \mathbb{R}^{13}$. 
Task completion is defined by a final position error of less than $5cm$ vertically and $10cm$ horizontally from the target landing point. 
% The task is successful when the agent’s Euclidean distance to the ground is within 10 cm, and its xy-plane distance to the landing point is within 20 cm.
\paragraph{Racing}
Autonomous drone racing, as one of the most popular tasks has garnered widespread attention. Racing requires the quadrotor to pass through a series of gates in a given order as fast as possible without losing control. We define the observation space as $\bm {s}_t^{\text{drone}} = \left[\bm {p}_t^{r_1},\bm {p}_t^{r_2}, \bm {v}_t,\bm {q}_{t}, \boldsymbol{\omega}_{t} \right] \in \mathbb{R}^{16}$. The $\bm {p}_t^{r_1}\in\mathbb{R}^{3}$ and $\bm {p}_t^{r_2}\in\mathbb{R}^{3}$ correspond to the relative position of the drone to the next two gates' centers. The task is successful when the agent traverses all gates in the correct order, thereby completing one full lap.

\subsubsection{Vision-based Tasks}
\paragraph{Visual Landing}
The visual landing task requires the quadrotor to land using vision input rather than the target's position. The observation space include the semantic segmentation and basic state defined by $\bm {s}_t^{\text{semantic}} \in \mathbb{R}^{64 \times 64}$ and $\bm {s}_t^{\text{drone}} = \left[\bm {p}_t, \bm {v}_t, \bm {q}_{t}, \boldsymbol{\omega}_{t} \right] \in \mathbb{R}^{13}$. It is worth noting that this task does not require any target information or depth, which relies solely on the segmentation. During the training phase, we use the landing platform measuring $50cm \times 50cm \times 20cm$ with a random shape (circle, square, triangle). The task completion is the same as the state-based landing task.

% To facilitate training, we designed the apron as a $50cm \times 50cm$ Apriltag QR code. 
% For the real-world experiment, we used a downward-facing monocular camera costing less than 15\$ to provide a stable 30 Hz RGB input. 
% , where $\bm {p}_t$ represent the relative position to the landing point.
\paragraph{Racing in Cluttered Environment}
Another vision-based task involves racing with random obstacles, which requires perceiving external obstacles from visual input. Therefore, the observation space includes both depth map and basic state defined by $\bm {s}_t^{\text{depth}}\in \mathbb{R}^{64\times64}$ and $\bm {s}_t^{\text{drone}} =\left[\bm {p}_t^{1},\bm {p}_t^{2}, \bm {v}_t, {v}_d, \bm {q}_{t}, \boldsymbol{\omega}_{t}\right]$. Task completion is defined as flying one full lap without collisions.

Specifically, our end-to-end policy is trained to map observations directly to CTBR commands. The action is defined by $a_t=[\bm T_t,\boldsymbol{\omega}^{x}_t,\boldsymbol{\omega}^{y}_t,\boldsymbol{\omega}^{z}_t]\in \mathbb{R}^{4}$, where $\bm T_t$ and $\boldsymbol{\omega}^{x}_t,\boldsymbol{\omega}^{y}_t,\boldsymbol{\omega}^{z}_t$ response the collective thrust and bodyrate respectively. We clamp the action to $[-1, 1]$ during training and use the ReLU activation function at the policy network's last layer to keep the policy outputs within a fixed range.

For the reward functions, we follow the provided instruction in Sec. \ref{reward_mannul}. In Sec. \ref{diff_reward}, we compare how different reward formulations affect the performance of the same task.

\subsection{Policy Training}\label{policy_training}
To verify that our system can perform robust and efficient zero-shot transfer across various algorithms, we employ BPTT \cite{ABPT} and PPO \cite{ppo} as the representatives of differentiable simulation and RL for policy training. The parameters for both algorithms are shown in Table \ref{table:training_para}. All experiments are run on a 32-core 13th Gen Intel(R) Core(TM) i9-13900K processor and an RTX-4090 GPU. 

% For all state-based benchmarks, the tasks are trained for $5\times10^7$ steps, except for the racing task, which requires $1\times10^8$ steps. In vision-based tasks, the landing task converged after $1\times10^6$ steps with BPTT, whereas the perception-aware racing task required $1\times10^8$ time steps with PPO.

\begin{table}[h]
\centering
\caption{Parameters for policy training.}
\label{table:training_para}
\begin{tabular}{lll}
\toprule
 & \textbf{Parameter} & \textbf{Value} \\
\midrule
\textbf{BPTT} & optimizer & Adam \\
 & learning rate & 1e-3 decay to 1e-5 \\
 & discount factor  & 0.99 \\
 & horizon length  & 96 \\
 & replay buffer size & 100000 \\
& batch size & 25600 \\
 & number of parallel environments & 100 \\
\midrule
\textbf{PPO} & optimizer & Adam \\
 & learning rate & 1e-4 decay to 1e-5 \\
 & discount factor  & 0.99 \\
 & clip range  & 0.2 \\
 & GAE-$\lambda$  & 0.95 \\
 & batch size & 25600 \\
 & number of parallel environments & 100 \\
\midrule
\end{tabular}
\end{table}

\subsection{Learning via Reinforcement Learning and Differentiable Simulation}\label{learning_rl_ds} % 考虑还要不要加别的RL和DS算法，或者针对连续和不连续的奖励函数情况分析两种算法的优劣势？

\begin{table*}[htbp]
\centering 
\caption{The observation space and reward function of state-based tasks.}
\label{tab:reard_rl_and_ds} 
\begin{tabular}{cccc} 
\toprule 
\textbf{Task} & \textbf{Observation} & \textbf{Reward} & \textbf{Example Parameters} \\ \midrule
Hovering & $(\bm p_t,\bm q_t,\bm v_t,\bm a_t)\in \mathbb{R}^{13}$ & $r_{t}^{\text{prog2}}+r_{t}^{\text{ori2}}+r_{t}^{\text{v1}}+r_{t}^{\text{anglev1}}$(D) & $\lambda_2=0.01,\lambda_{10}=0.0001,\lambda_{13}=0.002,\lambda_{17}=0.002$\\
Landing & $(\bm p_t,\bm q_t,\bm v_t,\bm a_t)\in \mathbb{R}^{13}$ &$r_t^{\text{progxy}}+r_{t}^{\text{vz}}+r_t^{\text{anglev1}}$(D)  &$\lambda_3=0.04,\lambda_{16}=0.1,\lambda_{17}=0.001$\\
Tracking & $(\bm p_t^{i=1-10},\bm q_t,\bm v_t,\bm a_t)\in \mathbb{R}^{40}$ &$r_{t}^{\text{prog2}}+r_{t}^{\text{ori2}}+r_{t}^{\text{v1}}+r_{t}^{\text{anglev1}}$(D) &$\lambda_2=0.02,\lambda_{10}=0.001,\lambda_{13}=0.002,\lambda_{17}=0.002$ \\
Racing & $(\bm p_t^1,\bm p_t^2,\bm q_t,\bm v_t,\bm a_t)\in \mathbb{R}^{16}$ & $r_{t}^{\text{prog1}}+r_{t}^{\text{ori2}}+r_{t}^{\text{v1}}+r_{t}^{\text{anglev1}}$(D) & $\lambda_1=0.9,\lambda_{10}=0.001,\lambda_{13}=0.002,\lambda_{17}=0.002$\\
\bottomrule 
\end{tabular}
\begin{tablenotes}
  \item[a] The table above provides the reward function designs for state-based tasks in this paper, with each reward term derived from the reward function manual in the methodology section. Here, D denotes fully differentiable rewards.\( \bm p \), \(\bm q \), \(\bm v \), and \(\bm a \) represent the relative position between the body center and the target, the orientation expressed in quaternions, the linear velocity, and the angular velocity, respectively.
\end{tablenotes}
\end{table*}

This paper presents the first comprehensive, task-by-task experimental comparison between RL and differentiable simulation. Four classical state-based tasks, hovering, landing, tracking, and racing, are used as benchmarks. To ensure fairness, both PPO and BPTT share identical reward functions, observation spaces, network architectures, training parameters, and policy outputs. The observation definitions and reward structures are summarized in Table~\ref{tab:reard_rl_and_ds}, while the reward curves and training efficiency metrics are presented in Fig.~\ref{fig:diff_method_reward} and Table~\ref{table:training}.

Across all tasks, Fig.~\ref{fig:diff_method_reward} shows that BPTT achieves substantially faster convergence and consistently reaches higher reward values than PPO. As detailed in Table~\ref{table:training}, the difference in FPS between the two methods is minimal, indicating comparable per-step computational cost. However, BPTT requires significantly fewer environment steps to converge, resulting in markedly shorter overall training times. On average, BPTT converges in approximately 1,200 seconds, representing less than 30\% of PPO’s training time of roughly 4,150 seconds. This superior efficiency can be attributed to the availability of precise analytical gradients in differentiable simulation. With an accurate quadrotor dynamics model, BPTT directly obtains the gradient of the policy objective with respect to the parameters, enabling rapid and stable optimization. As a result, only a small number of iterations are needed to reach high-performing policies, demonstrating the high sample efficiency of the differentiable simulation. In contrast, PPO relies on stochastic gradient estimates derived from extensive sampling, which leads to slower convergence despite eventually producing policies capable of accomplishing each task. The simulation results of the trained policies are illustrated in Fig.~\ref{fig:task_sim}, confirming the effectiveness of both methods and highlighting the significantly higher training efficiency of BPTT. 

\subsection{Learning under Different Reward Functions}\label{diff_reward} 
% To demonstrate that the proposed reward function designs enable diverse approaches to reward formulation for task completion, we compare task performance under various reward configurations. We adopt four state-based tasks as benchmarks and employ both BPTT and PPO to evaluate how these tasks perform when configured with the reward combinations specified in Sec. \ref{reward_mannul}. Specifically, for each task, we devise various reward functions that independently examine (1) the effectiveness of proposed design manual for specific tasks and (2) the presence or absence of each reward component proposed in Sec. \ref{reward_mannul}. The task evaluation curves for both PPO and BPTT are shown in Fig. \ref{fig:diff_reward_sr} and Fig. \ref{fig:diff_reward_sr_ds}. Task performance is measured by success rate in the landing and racing, and by positional error in the hovering and tracking.
\begin{figure*}[htbp]
    \centering
    \includegraphics[width=1.0\linewidth]{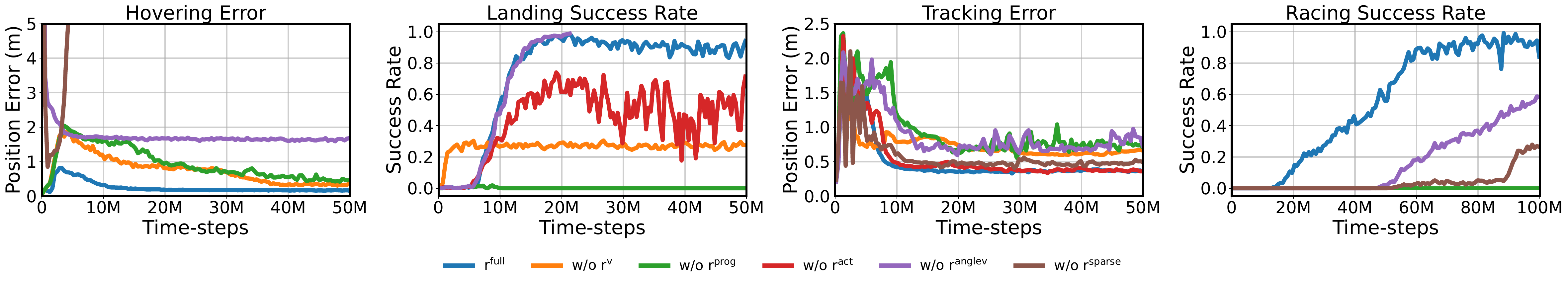}
    \caption{\textbf{Success rate and task error under different reward settings training via PPO.} The figure presents the ablation study evaluating the effects of the individual reward components trained with PPO. The $r^{\text{full}}$ represents the configuration using all reward components listed in Table \ref{tab:reward_tasks}, while w/o $r^{\text{v}}$,w/o $r^{\text{prog}}$,w/o $r^{\text{act}}$,w/o $r^{\text{anglev}}$,w/o $r^{\text{sparse}}$ denote the cases without linear-velocity reward, progress reward, action-smoothness reward, angular-velocity reward, and sparse reward, respectively. For the racing task, only the $r^{\text{full}}$, w/o $r^{\text{prog}}$, w/o $r^{\text{sparse}}$, w/o $r^{\text{anglev}}$ reward curves are included.}
    \label{fig:diff_reward_sr}
\end{figure*}

\begin{figure*}[htbp]
    \centering
    \includegraphics[width=1.0\linewidth]{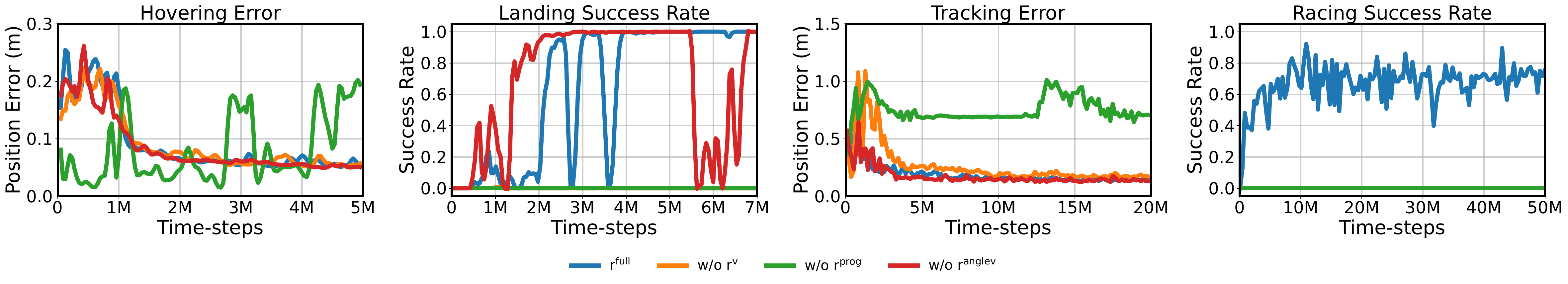}
    \caption{\textbf{Success rate and task error under different reward settings training via BPTT.} The figure presents the ablation study evaluating the effects of the individual reward components trained with BPTT. The $r^{\text{full}}$ represents the configuration using all reward components listed in Table \ref{tab:reward_tasks}, while w/o $r^{\text{v}}$,w/o $r^{\text{prog}}$,w/o $r^{\text{anglev}}$ denote the cases without linear-velocity reward, progress reward, action-smoothness reward, angular-velocity reward, and sparse reward, respectively. For the racing task, only the $r^{\text{full}}$ and w/o $r^{\text{prog}}$ reward curves are included.}
    \label{fig:diff_reward_sr_ds}
\end{figure*}

To demonstrate that the reward-design guidelines proposed in Sec. \ref{reward_mannul} can effectively accomplish different tasks and to validate the contribution of each reward component, we compare task performance under various reward configurations. Four state-based tasks are selected as benchmarks, and both BPTT and PPO are employed to evaluate performance under the specified reward combinations. Specifically, for each task, we design multiple reward functions to independently examine (1) the effectiveness of the proposed reward design instruction for the given task, and (2) the influence of including or excluding each reward component introduced in Sec. \ref{reward_mannul} on the overall task performance. The task evaluation curves for PPO and BPTT are shown in Fig. \ref{fig:diff_reward_sr} and Fig. \ref{fig:diff_reward_sr_ds}, respectively. The evaluation metrics include success rates for landing and racing, as well as position error for hovering and tracking.

As shown in Fig. \ref{fig:diff_reward_sr} and Fig. \ref{fig:diff_reward_sr_ds}, the rewards designed following our proposed instruction effectively guide both PPO and BPTT to accomplish various state-based tasks. The fundamental reward components, including progress, action smoothness, linear velocity, and angular velocity, are essential for both BPTT and PPO in stability-oriented tasks such as hovering, landing, and tracking. The absence of any component degrades task performance. These elements work together to maintain drone stability while completing tasks, preventing loss of control at high speeds or under large attitude angles. For the landing task, distinct progress rewards for the xy-plane and the z-axis are introduced to accommodate their respective velocity constraints. In racing, where precise attitude control and stabilization demands are relaxed, a heavily weighted progress reward plays a dominant role in accomplishing the task.

Furthermore, terminal rewards provide critical guidance for RL algorithms such as PPO. The sparse rewards issued upon task completion, along with collision penalties, enable the agent to recognize the influence of specific behaviors on long-term cumulative returns, thereby steering policy optimization toward successful task completion and crash avoidance. Overall, the experimental results further validate that our general-purpose reward-design methodology reduces tuning effort and remains effective across different tasks and algorithms. By systematically selecting and combining appropriate reward components according to specific task requirements, our framework streamlines the training process and ensures efficient policy convergence.

\begin{figure*}[htbp]
    \centering
    \includegraphics[width=0.98\linewidth]{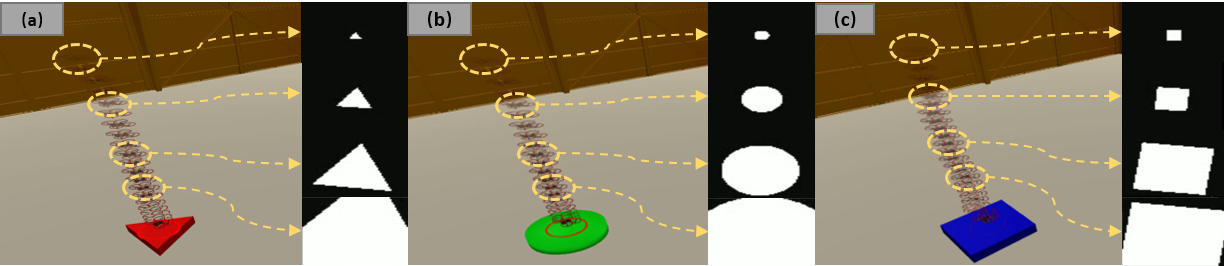}
    \caption{\textbf{Simulation result of vision-based landing.} Subfigures (a), (b), and (c) depict the performance of the identical policy while landing on triangular, circular, and square landing pads, respectively. The corresponding segmentation maps from the downward-facing camera are displayed on the right side of each subfigure.}
    \label{fig:diff_land}
\end{figure*}

\begin{figure*}[htbp]
    \centering
    \includegraphics[width=0.98\linewidth]{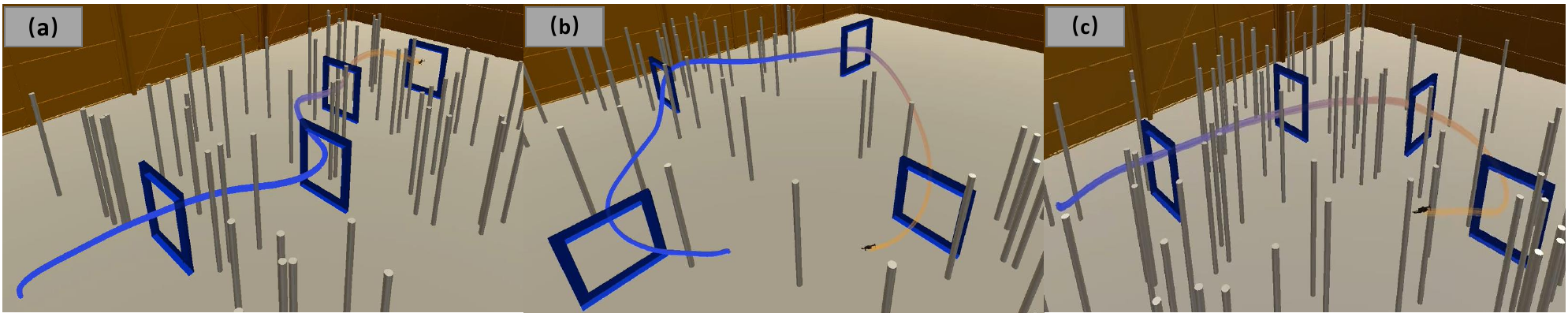}
    \caption{\textbf{Simulation result of racing with obstacles.} Subfigures (a), (b), and (c) present simulation results obtained in S-shaped, 3D circle, and J-shaped race tracks, respectively, with maximum velocities achieving over 10 m/s.}
    \label{fig:vision_racing}
\end{figure*}

\subsection{Learning with Visual Input}\label{learning_visual}
\begin{table*}[htbp]
\centering 
\caption{The observation space and reward function of vision-based tasks.}
\label{tab:reward_visual} 
\begin{tabular}{lcc} 
\toprule 
 & \textbf{Visual landing} & \textbf{Racing with obstacles} \\ 
\midrule
\textbf{Observation} & $(\bm p_t,\bm q_t,\bm v_t,\bm a_t)\in \mathbb{R}^{13}$ \& $ {segmentation} \in \mathbb{R}^{64 \times 64}$ & $(\bm p_t^1,\bm p_t^2,\bm q_t,\bm v_t,v_d,\bm a_t)\in \mathbb{R}^{17}$ \& $ depth \in \mathbb{R}^{64 \times 64}$ \\ [0.3em] \midrule
\textbf{Reward} & $r_{t}^{\text{progxy}}+r_{t}^{\text{ori2}}+r_{t}^{\text{vz}}+r_{t}^{\text{anglev1}}$(D) & $r_{t}^{\text{prog1}}+r_{t}^{\text{act3}}+r_{t}^{\text{ori1}}+r_{t}^{\text{avoid1}}+r_{t}^{\text{goal}}+r_{t}^{\text{crash}}$(PD) \\[0.3em]\midrule
\textbf{Parameter} &$\lambda_3=0.04,\lambda_{10}=0.003,\lambda_{16}=0.1,\lambda_{17}=0.001$ & $\lambda_1=0.9,\lambda_7=0.025,\lambda_8=0.002,\lambda_9=0.04,\lambda_{18}=0.01,\lambda_{21}=5,\lambda_{22}=4$ \\[0.3em]
\bottomrule 
\end{tabular}
\begin{tablenotes}
  \item[a] The table above provides the reward function designs for vision-based tasks in this paper, with each reward term derived from the reward function manual in the methodology section. Here, D denotes fully differentiable rewards, while PD denotes partially differentiable rewards. \( \bm p \), \(\bm q \), \(\bm v \), and \(\bm a \) represent the relative position between the body center and the target, the orientation expressed in quaternions, the linear velocity, and the angular velocity, respectively.
\end{tablenotes}
\end{table*}

\begin{figure*}[htbp]
    \centering
    \includegraphics[width=1.0\linewidth]{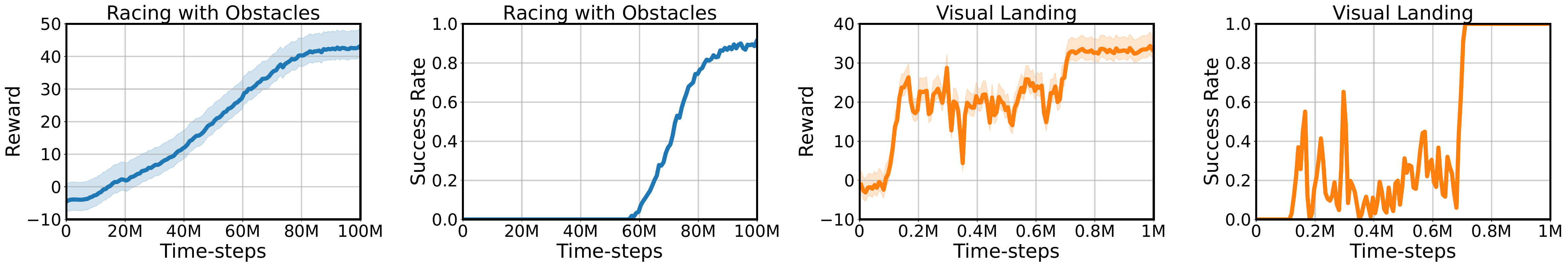}
    \caption{\textbf{Reward and success rate curves of visual tasks.} From left to right, the figure depicts the reward and success rate curves for racing with obstacles and visual landing. }
    \label{fig:reward_vision}
\end{figure*}

% To validate the effectiveness of our system for vision-based tasks, we use landing with segmentation and racing in cluttered environments with a depth map as benchmarks for this section. To validate the compatibility of our system with visual inputs across diverse algorithms, the landing and racing tasks are trained by BPTT and PPO, respectively. The reward and success rate curves are shown in Fig. \ref{fig:reward_vision}. The simulation results of both tasks are shown in Fig. \ref{fig:diff_land} and Fig. \ref{fig:vision_racing}.
To assess the effectiveness of our framework on vision-based tasks, we evaluate two representative benchmarks: landing with semantic segmentation and obstacle-aware racing using depth maps. To demonstrate compatibility with diverse learning paradigms, we train the landing task using BPTT and the racing task using PPO. The reward curves and success rates are presented in Fig.~\ref{fig:reward_vision}, and the corresponding simulation results are shown in Fig.~\ref{fig:diff_land} and Fig.~\ref{fig:vision_racing}.

As illustrated in Fig.~\ref{fig:reward_vision}, our system enables efficient learning across both tasks and supports different algorithms without task-specific modifications. In the obstacle-aware racing scenario, PPO combined with a curriculum learning strategy successfully generalizes to unseen environments and achieves a 100\% success rate across three racetracks of different shapes, as shown in Fig.~\ref{fig:vision_racing}. For the vision-based landing task, BPTT achieves end-to-end single-step training using only the semantic segmentation map, without any additional prior knowledge of the landing target. The trained policy exhibits strong generalization across diverse initial positions and landing platforms with varying appearances, as illustrated in Fig.~\ref{fig:diff_land}. Notably, BPTT requires only $1\times10^6$ time steps to complete the landing task, highlighting the substantial sample efficiency enabled by analytical gradient back-propagation. Together, these two vision-based tasks demonstrate the effectiveness of our system in learning robust policies from visual observations.

% As observed in Fig. \ref{fig:reward_vision}, our system achieves rapid training and task completion in vision-based tasks, while demonstrating adaptability to different algorithms. For racing with obstacles, by incorporating a depth map and adopting a curriculum learning strategy, the policy trained by PPO successfully generalizes to unseen obstacle-aware environments, as shown in Fig. \ref{vision_racing}. For vision-based landing, we have implemented end-to-end policy training without any target observations using BPTT. As shown in Fig. \ref{fig:diff_land}, the trained policy generalizes to different initial positions and even landing platforms of varying shapes, further highlighting our system's strong robustness to visual observations and algorithmic choices. It is worth noting that the differentiable simulation approach requires only $1e6$ time steps for training in this task.

\begin{figure*}[htbp]
    \centering
    \includegraphics[width=1.0\linewidth]{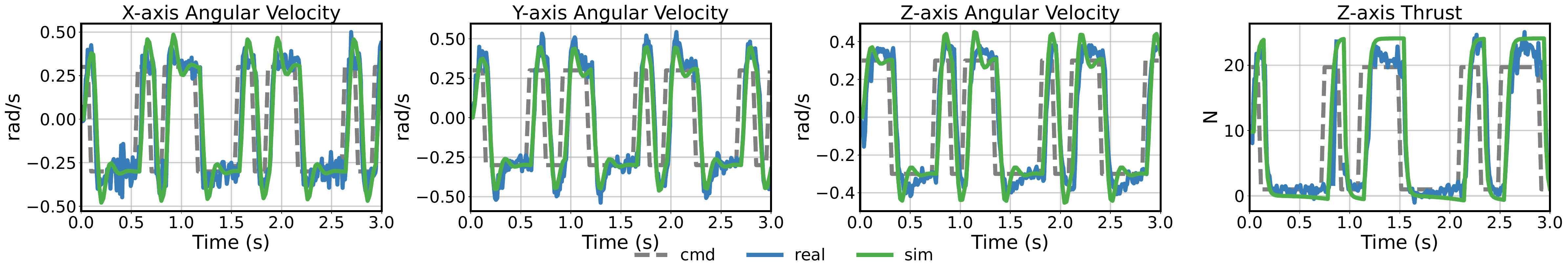}
    \caption{\textbf{The step-response alignment.} We align the step-response of angular velocity and the mass-normalized thrust along the z-axis.}
    \label{fig:step_response}
\end{figure*}
\begin{figure*}[htbp]
    \centering
    \includegraphics[width=1.0\linewidth]{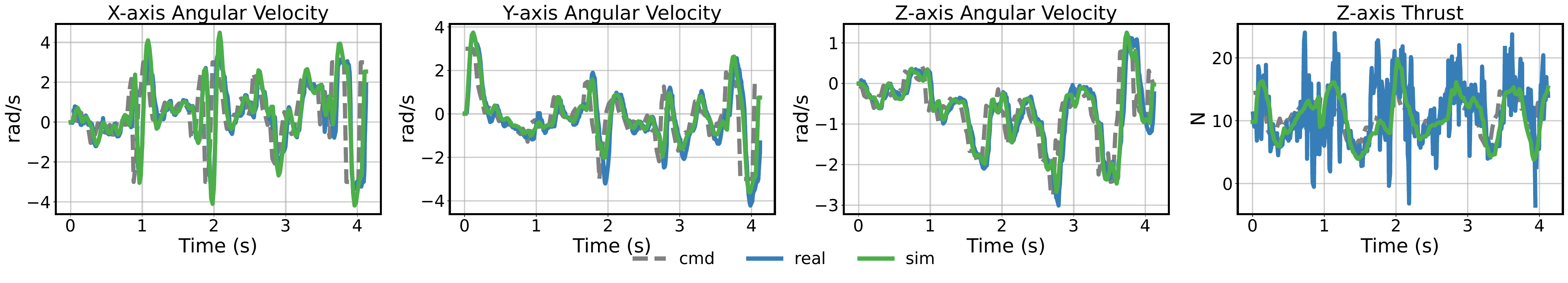}
    \caption{\textbf{Policy response after step-response alignment.}  The corresponding policy deployment without requiring additional alignment, yielding a response that closely matches the simulation.}
    \label{fig:policy_response}
\end{figure*}

\subsection{Zero-shot sim-to-real transfer}
\begin{figure*}[htbp]
    \centering
    \includegraphics[width=1.0\linewidth]{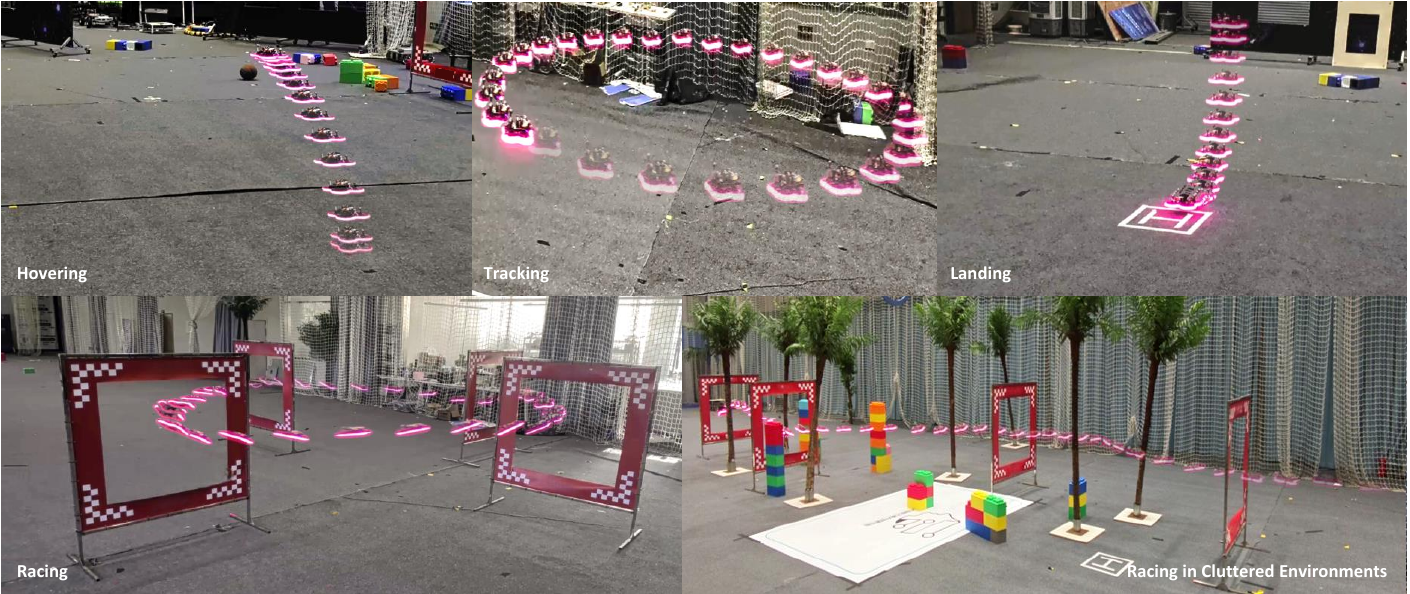}
    \caption{\textbf{Zero-shot transfer from simulation to real world via our E2E-Fly system.} Real-world performance demonstrates that the policy trained with our system can achieve zero-shot transfer from simulation to the real world.}
    \label{fig:task_real}
\end{figure*}

To validate the effectiveness of the proposed sim-to-real alignment methodology, we begin by analyzing the step response alignment, as illustrated in Fig. \ref{fig:step_response}. This step is essential because discrepancies in system dynamics or delay characteristics directly translate into inconsistent behavior. Subsequently, we demonstrate the performance of a fully aligned policy in both simulation and the real world (Fig. \ref{fig:policy_response}). All real-world flight experiments are conducted on the VIS-R platform, while the aggressive task, racing with random obstacles, is additionally validated through hardware-in-the-loop simulation on the VIS-H platform. Finally, to verify whether our system enables zero-shot policy transfer, we evaluate each policy in simulated and real-world experiments, with corresponding results presented in Fig. \ref{fig:task_sim} and Fig. \ref{fig:task_real}, respectively.

As shown in Fig.~\ref{fig:step_response} and Fig.~\ref{fig:policy_response}, the combination of system identification, latency compensation, domain randomization, and noise modeling produces a close match between simulated and real-world response curves. This alignment enables reliable zero-shot deployment without any task-specific system parameters fine-tuning. Practically, this means that once a policy has converged in simulation, it can be executed directly on the physical platform while preserving its functional characteristics. This capability underscores the critical importance of comprehensive sim-to-real alignment in quadrotor research. The task-level comparisons in Fig.~\ref{fig:task_sim} and Fig.~\ref{fig:task_real} further confirm that the learned policies exhibit nearly equivalent performance across simulated and real environments, demonstrating the robustness and effectiveness of the proposed sim-to-real alignment pipeline.

\section{Discussion}\label{dis}

\subsection{Difficulty for Differentiable Simulation to Design Reward}

Despite its theoretical advantages, BPTT in differentiable simulation faces significant challenges when applied to complex, long-horizon tasks. A primary limitation lies in the difficulty of designing effective and smoothly differentiable reward functions. Inappropriate reward landscapes can cause gradient instability during back-propagation through extended temporal sequences, leading to vanishing or exploding gradients that hinder stable policy optimization. Furthermore, as gradient optimization inherently performs a local search, it is prone to convergence toward suboptimal behaviors driven by local reward gradients rather than global task objectives. For example, in a landing task, an excessively steep differentiable penalty on deviation from the target may bias the drone toward a slow, conservative descent, while obscuring the path to a more agile, time-optimal solution.

\subsection{Combining the Advantages of Differentiable Simulation and RL}
RL does not require knowledge of the precise environment model. Instead, the agent learns policies from reward signals through interaction with the environment. Even with specific dynamic differences between simulation and the real world, techniques such as domain randomization can improve the policy’s robustness. Thanks to its strong exploratory capabilities, RL may discover unconventional yet efficient behavioral patterns. However, it suffers from low sample efficiency and unstable training.

Differentiable simulation, on the other hand, leverages precise differentiable dynamics to optimize policies via gradient back-propagation directly. It boasts extremely high sample efficiency. Nevertheless, BPTT has drawbacks: it not only requires constructing a fully differentiable gradient chain but also makes reward function gradients highly sensitive, posing significant challenges for learning from sparse rewards. 

Therefore, RL is well-suited for complex problems with unknown models that require exploration, while differentiable simulation is ideal for tasks with precise, differentiable models that demand efficient, accurate control. However, the two are not mutually exclusive but highly complementary. We believe the future trend is to combine their strengths: using RL's exploratory power to learn high-level goals and employing differentiable simulation for precise local motion and control optimization at the lower level.
% \subsection{Difficulty in generalizing from one policy to multi-tasks}
% End-to-end learning policies have demonstrated good performance in single tasks and successfully achieved transfer from simulation to reality. This success is attributed to both the design of task-specific reward functions and the effective use of training tricks. Although these implementation details have been elaborated in our system, most existing methods are only adapted to a single task objective, and one specific policy often struggles to handle multi-task objectives simultaneously. Moreover, the generalization of a single policy to tasks with different observations and rewards is relatively poor, and most methods are not directly end-to-end trained. Therefore, end-to-end learning-based methods for more complex multi-task scenarios remain a challenging problem that needs to be addressed.

\section{Conclusion}
In conclusion, we present E2E-Fly, a unified framework that bridges the sim-to-real gap for quadrotors via an integrated pipeline composed of policy training, validation, and real-world deployment. By synergistically combining differentiable physical learning in a high-speed rendering environment and supporting it with structured reward design and curriculum learning, our system enables efficient policy optimization for diverse state-based and vision-based tasks. The framework ensures robust policy transfer through a comprehensive validation strategy involving sim-to-sim evaluation and hardware-in-the-loop test, and a systematic sim-to-real alignment methodology that includes system identification, latency compensation, domain randomization, and noise modeling. Extensive experiments across six challenging tasks confirm that our approach achieves reliable zero-shot transfer, deploying trained policies directly to physical quadrotors with high performance, thereby offering a complete and reproducible platform for end-to-end learning in agile autonomous flight. Our work establishes a complete pipeline and an efficient sim-to-real deployment framework for learning-based end-to-end algorithms in quadrotors, laying a solid platform foundation for future research.

\bibliographystyle{IEEEtran}
% \bstctlcite{IEEEexample:BSTcontrol} 
% 参考格式根据不同刊物要求进行更改即可。
% 如IEEEtran, plain, unsrt， alpha, abbrv, ieeetr, acm, siam等等。
\bibliography{Reference}

% \vfill

\end{document}